\documentclass[final]{arxiv}
\nolinenumbers

\DOI{xxxxx}
\Year{2026}
\usepackage{lipsum}
\usepackage{siunitx}
\usepackage{fancyvrb}
\usepackage{fvextra}
\usepackage{booktabs} 
\usepackage{subcaption}
\usepackage{xspace}
\usepackage{comment}
\usepackage{enumitem}
\usepackage{xcolor}
\usepackage{float}
\usepackage{makecell}
\usepackage{multirow}
\usepackage[skip=5pt]{parskip} 
\DeclareSIUnit{\pp}{\%pt}

\definecolor{navyblue}{RGB}{0,0,128}
\usepackage[
    colorlinks=true,
    linkcolor=navyblue,
    citecolor=navyblue,
    urlcolor=navyblue
]{hyperref}

\makeatletter
\def\authormark#1{}
\makeatother

\newcommand{\refSection}[1]{Section~\nameref{#1}{}}

\acrodef{UAV}{Uncrewed Aerial Vehicle}
\acrodef{GSD}{Ground Sampling Distance}
\acrodef{ALS}{Airborne Laser Scanning}
\acrodef{CNN}{Convolution Neural Network}
\acrodef{AI}{Artificial Intelligence}
\acrodef{SAM}{Segment Anything Model}
\acrodef{Grad-CAM}{Gradient-weighted Class Activation Mapping}
\acrodef{GAN}{Generative Adversarial Network}
\acrodef{LLM}{Large Language Model}
\acrodef{FID}{Fréchet Inception Distance}
\acrodef{CMMD}{CLIP Maximum Mean Discrepancy}
\acrodef{mIoU}{mean Intersection over Union}
\acrodef{mIPQ}{mean Isoperimetric Inequality Quotient}
\acrodef{LLM}{Large Language Model}
\acrodef{VLM}{Vision Language Model}
\acrodef{ViT}{Vision Transformer}
\acrodef{VFM}{Vision Foundation Model}
\acrodef{IQR}{Interquartile Range}
\acrodef{API}{Application Programming Interface}

\newcommand{\nasiridataset}{\mbox{\textsc{WilDReF-Q}}\xspace}
\newcommand{\ourdataset}{\mbox{\textsc{WilDReF-Q-V2}}\xspace}
\newcommand{\GenAiDataset}{\textsc{Gen4Regen}\xspace}
\newcommand{\GenAiDatasetName}{\textsc{Gen4Regen}}  

\newcommand{\numlabelled}{\num{199}\xspace}
\newcommand{\numunlabelled}{\num{25106}\xspace}
\newcommand{\numaddedlabelled}{\num{50}\xspace}
\newcommand{\numaddedunlabelled}{\num{13977}\xspace}

\newcommand{\numregen}{\num{2101}\xspace}
\newcommand{\numinaturalist}{\num{318151}\xspace}
\newcommand{\numclasses}{\num{23}\xspace}
\newcommand{\numplantclasses}{\num{20}\xspace}
\newcommand{\numnonplantclasses}{three\xspace}
\newcommand{\numbioclimdomains}{four\xspace}
\newcommand{\numbioclimdomainsnewdata}{three\xspace}
\newcommand{\numsitestotal}{\num{12}\xspace}

\let\cite\citep

\begin{document}

\vspace*{-3\baselineskip}
\title[{\normalfont The \textsc{Gen4Regen} Dataset for Forest Regeneration Mapping}]{Leveraging Image Generators to Address Data Scarcity: \\ The \textsc{Gen4Regen} Dataset for Forest Regeneration Mapping}

\author[{\normalfont Leveraging Image Generators to Address Data Scarcity}]{Gabriel \surname{Jeanson}$^{1,2,\ast}$, David-Alexandre \surname{Duclos}$^{1,2}$, William \surname{Larrivée-Hardy}$^{1,2}$, Noé \surname{Cochet}$^{1,2}$,\\Matěj \surname{Boxan}$^{1,2}$, Anthony \surname{Deschênes}$^{2}$, François \surname{Pomerleau}$^{1,2}$, and Philippe \surname{Giguère}$^{1,2,\ast}$}

\address{
    $^{1}$Northern Robotics Laboratory, Université Laval, Québec, QC, G1V 0A6, Canada\\
    $^{2}$Département d'informatique et de génie logiciel, Université Laval, Québec, QC, G1V 0A6, Canada\\
}

\corres{$^{\ast}$Corresponding authors. E-mails: gabriel.jeanson@norlab.ulaval.ca, philippe.giguere@ift.ulaval.ca}

\begin{abstract}
\begin{spacing}{1.0}

Sustainable forest management relies on precise species composition mapping, yet traditional ground surveys are labour-intensive and geographically constrained. 
While \acp{UAV} offer scalable data collection, the transition to deep learning-based interpretation is bottlenecked by the severe scarcity of expert-annotated imagery, particularly in complex, visually heterogeneous regeneration zones.
This paper addresses the dual challenges of data scarcity and extreme class imbalance in the fine-grained semantic segmentation of plants by providing a scalable framework that reduces reliance on manual photo-interpretation for high-resolution, millimetre-level aerial imagery.
Importantly, we leverage the large-scale Nano Banana Pro model to simultaneously generate high-fidelity images and their corresponding pixel-aligned semantic masks from prompts.
We introduce \ourdataset, an expansion of a natural forest dataset with \numaddedunlabelled new unlabelled and \numaddedlabelled hand-labelled \textit{real} images, as well as the \GenAiDataset dataset, featuring \numregen pairs of \textit{synthetic} images and semantic masks.
Our methodology integrates real-world data with AI-generated images, addressing both data scarcity and class imbalance in forest regeneration mapping.
We evaluate the orthogonality of manual labels, automated pseudo-labels, and synthetic data using the Mask2Former, DINOv2, and DINOv3 architectures and highlight that AI-generated data is highly complementary to real-world pseudo-labels, with unified training yielding an F1 score improvement of over \SI{15}{\pp} compared to purely supervised baselines.
Furthermore, we demonstrate that even small quantities of prompt-generated data significantly improve performance for underrepresented classes, some of which see per-class F1 score gains of over \SI{30}{\pp}.
We conclude that large-scale vision models can serve as agile data generators, effectively bootstrapping perception tasks for niche AI domains where expert labels are scarce or unavailable.
This paradigm enables practitioners to bypass seasonal data collection constraints and develop specialized ecological monitoring systems with limited manual effort.
Our datasets, source code, and models will be available at \url{https://norlab-ulaval.github.io/gen4regen}.

\end{spacing}

\end{abstract}

\maketitle

\vspace{-24pt}
\section{Introduction}

\acresetall

Forest regeneration is critical for sustainable land management and biodiversity conservation, particularly following disturbances such as wildfire, disease, or logging.
Effective silvicultural operations play a crucial role in forest regeneration, especially with changing climate conditions \cite{routa_effects_2019, achim_changing_2022}.
However, the success of these operations depends on precise monitoring, including assessing restoration success and optimizing interventions such as planting, thinning, and plant control.
An important challenge of this forest regeneration surveillance lies in the accurate identification of species composition, which is crucial for detecting competing vegetation, dead trees, and debris.
These components are essential for supporting informed and timely management decisions.
Traditionally, these data were gathered via ground-based surveys, but their time-intensive nature and the logistical challenges of accessing remote, rugged terrain make them inefficient for surveying large forest areas.
Consequently, alternative approaches have received increasing attention, as reflected in a recent survey of over 300 forest practitioners highlighting strong interest in species composition mapping using remote sensing for silvicultural planning \cite{fassnacht_forest_2025}.

\begin{figure}[!h]
    \centering{\includegraphics[width=0.99\linewidth]{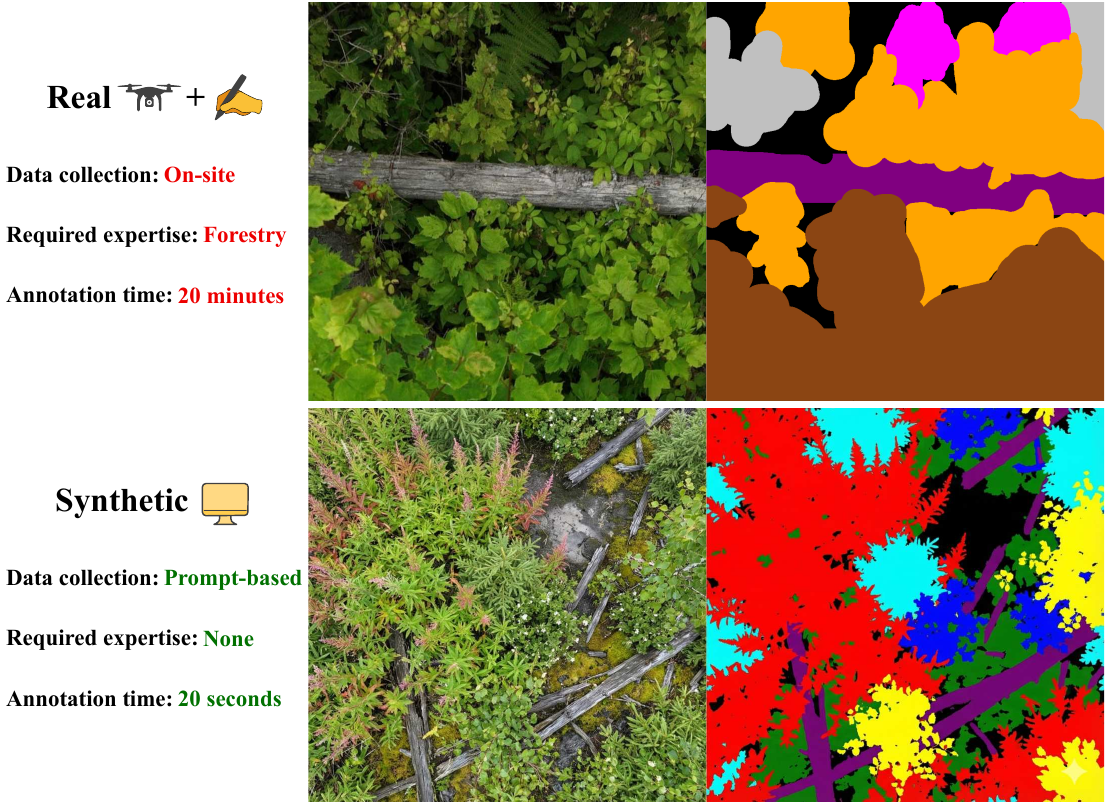}}
    \caption{Real \ac{UAV} image (top left) with its corresponding manually labelled semantic segmentation mask (top right), compared to a prompt-generated image (bottom left) and its associated prompt-generated mask (bottom right) produced using the large-scale Nano Banana Pro model.
    Real \ac{UAV} datasets in forestry require on-site data collection and forestry experts for annotation, resulting in long, costly, and coarse annotations.
    In contrast, AI image generators can now generate semantic masks paired with images with a high level of photorealism.
    }
    \label{fig:ai_vs_real_images}
\end{figure}

The transition to \ac{UAV} imagery provides a highly scalable alternative for data collection, but it fundamentally shifts the bottleneck from \textit{physical field access} to \textit{data interpretation}.
Deep learning is a promising approach to the latter, but supervised learning algorithms require large quantities of precisely labelled data to generalize effectively across environments.
While the broader machine learning community has directed significant resources to create datasets for mainstream perception tasks such as urban autonomous driving or robotic grasping, fine-grained ecological perception remains an orphaned \ac{AI} problem \cite{reynolds_potential_2025}.
Thus, from a computer vision perspective, our species composition mapping represents a data-scarce problem.
Moreover, the task requires models to navigate widespread class imbalance, visual heterogeneity, and fuzzy boundaries, particularly at fine spatial scales such as small grass leaves.
Unlike urban environments where bounding boxes are easily drawn by crowd-sourced workers, annotating high-resolution overhead images of vegetative cover requires \textit{specialized} photo-interpretation by forestry or botanical experts, resulting in a severe scarcity of real annotated images.
\autoref{fig:ai_vs_real_images} presents at the top an example of a \ac{UAV} image and its corresponding expert annotation. 
It highlights two problems: i) expert annotations are generally approximations, and ii) the labelling process is time-consuming, as every pixel of the high-resolution image needs an annotation.

Previous approaches for photo-interpretation with \ac{UAV} imagery have focused primarily on mature forests, using a traditional supervised learning approach relying on hand-labelled images.
For instance, for tree crown segmentation in natural environments, Detectree2 \cite{ball_accurate_2023} was trained with \SI{3.8}{k} annotations, while BAMFORESTS \cite{BAMFORESTS2024} relied on \SI{27}{k} annotations. 
Similar approaches have been done for tree species identification \cite{ferreira2020individual, ECKE2024108785}.
In all cases, these approaches require significant manual annotation efforts.
These limitations have increasingly motivated interest in alternative strategies that reduce reliance on exhaustively hand-labelled real‑world imagery.
In addition to the data scarcity problem, class imbalance also causes issues for deep learning solutions in natural forests.
One way to solve class imbalance in semantic segmentation, a key issue in our situation, is to perform random oversampling of minority classes. 
However, this approach is ill-suited to images with strong intra-image class imbalance, such as vegetative cover scenes.
To mitigate this limitation, some alternative approaches have been proposed.
For instance, \citet{soltani_automated_2025} present an automatic way of segmenting images of plants from one data source (i.e., citizen science data) and then pasting them into \ac{UAV} imagery.
However, the quality of the augmentation is limited by the ability to segment the plant accurately, and then to harmonize its appearance with the target domain (i.e., \ac{UAV} imagery).
An alternative approach to tackle both issues is the use of synthetic data \cite{grondin_tree_2023,blaga_forest_2026}.
These efforts have historically relied on explicit procedural modelling using 3D assets and game engines, and although they can achieve an impressive level of diversity and realism \cite{feng_spread_2025}, they still require experts and can struggle to accurately replicate the unstructured nature and diversity of forest regeneration environments that define our operating context.

In response to these considerations, we propose a comprehensive methodology that combines several techniques to address the lack of labelled data and class imbalance.
First, we make use of extremely small \ac{GSD}, down to the millimetre-level.
As recently demonstrated by \citet{soltani_simple_2024} and \citet{nasiri_using_2025}, leveraging ultra-low-altitude \ac{UAV} imagery captures intricate leaf structures comparable to citizen science images.
Such resolution effectively bridges the domain gap between aerial imagery and ground-level citizen science, enabling automated pseudo-labelling workflows that reduce the need for manual annotation.
To this effect, we introduce an extension of the \nasiridataset dataset for semantic segmentation of plants in Quebec forest regeneration zones presented in \cite{nasiri_using_2025}, featuring \numaddedlabelled new hand-labelled images and \numaddedunlabelled new unlabelled images taken in \numbioclimdomainsnewdata bioclimatic domains in the province of Quebec.
Following \citet{nasiri_using_2025}, we leverage citizen science data to produce pseudo-labels on the full unlabelled dataset.
Second, we explore the use of generative AI as a \textit{prompt-based training data generator}.
Recent findings by \citet{gabeur_image_2026} demonstrate that large-scale image generation pre-training plays a foundational role analogous to large language model pre-training, equipping models such as Google's Nano Banana Pro \cite{google_gemini_2026} with deep, internalized representations of visual semantics and underlying geometric structures.
Incidentally, \citet{zuo2025nanobananaprolowlevel} also demonstrate Nano Banana Pro's strong zero-shot capabilities for semantic tasks.
Consequently, we present a novel dataset called \GenAiDataset, comprising \numregen images and corresponding labels generated using the Nano Banana Pro model, as seen at the bottom of \autoref{fig:ai_vs_real_images}.
This approach allows practitioners in label-scarce domains, such as ours, to generate data for underrepresented classes under specific contexts via text prompts, further increasing domain diversity and improving generalization.
We also present actionable guidelines on harmonizing these generated image-mask pairs with real data to effectively mitigate the domain gap.
To the best of our knowledge, we are the first to demonstrate that large-scale image generators are now capable of jointly synthesizing high-fidelity, photorealistic images alongside their precise semantic masks, in a single inference step.

In short, our contributions are:
\begin{itemize}[topsep=0pt, labelsep=1em, leftmargin=*]
    \item \GenAiDataset, our novel synthetic dataset generated with Nano Banana Pro, with photorealistic images of forest regeneration environments and corresponding generated high-precision semantic segmentation masks;
    \item An assessment of bootstrapping data-scarce, class-imbalanced projects by analyzing the use of \GenAiDataset as a training dataset for semantic segmentation alongside other data sources; 
    \item \ourdataset, an extension of the \nasiridataset dataset with \ac{UAV} images of natural forest regeneration vegetation, featuring new hand-labelled and unlabelled images, an additional camera sensor, and improved annotations;
    \item An evaluation of zero-shot semantic segmentation for species composition mapping using Nano Banana Pro.
\end{itemize}

\section{Related work}

In this section, we first present an overview of supervised deep learning for \ac{UAV} plant classification, highlighting both its capabilities and the scalability issues caused by manual annotation.
Then, we will examine approaches for automated labelling leveraging citizen science data to generate pseudo-label semantic masks without requiring hand-labelled annotations.
These pseudo-labels will help establish a strong baseline for improving generalization capabilities.
Finally, we will discuss the use of synthetic data in deep learning, exploring the transition from traditional augmentations to advanced generative models to overcome data scarcity and class imbalance in complex environments.

\subsection{Supervised Deep Learning for \ac{UAV} Plant Classification}

Previous research has demonstrated the potential of deep learning methods for plant classification and segmentation using \ac{UAV} imagery.
While lidar, multispectral, and hyperspectral sensors achieve strong performance in forestry applications, their high cost remains a significant limitation \cite{chang_application_2025}.
In contrast, \acp{UAV} equipped with RGB cameras are more affordable, thus widely accessible to forestry practitioners.
They also offer higher spatial resolution compared to multispectral and hyperspectral sensors, which is of high importance for precise mapping \cite{Hao02102023, nasiri_using_2025}.
As such, several studies have investigated the use of deep learning with RGB \ac{UAV} imagery for forestry applications.
For example, \citet{schiefer_mapping_2020} used aerial RGB images for semantic segmentation of nine tree species, three genus-level classes, and two non-plant classes.
For this, they manually annotated \num{3112} tiles of $512 \times 512$ pixels to train a deep learning semantic segmentation model.
In another work, \citet{cloutier_influence_2024} studied the influence of leaf phenology on the instance segmentation of \num{11} tree species and two genera, using a dataset containing \num{23000} manually labelled tree crowns at \SI{1.6}{cm} spatial resolution.
They describe their demanding annotation process and point out how the quality of the annotations directly affects the quality of the predictions.
Despite the overall average F1 score of \SI{72}{\%} achieved with their supervised approach, their reliance on vast quantities of expert-annotated data creates a labelling scalability problem.
Similarly, \citet{chadwick_transferability_2024} performed instance segmentation of two tree species using high-resolution RGB images of \SI{3}{cm} \ac{GSD} in post-harvest forest regeneration zones.
The dataset used in this study contains a total of \num{403} images of $300 \times 300$ pixels: \num{162} ground-verified annotations used both for training and evaluation, as well as \num{241} photo-interpreted images used for training and validation.
More recently, \citet{duguay2026selvamask} introduced SelvaMask, a densely annotated tropical tree crown instance segmentation dataset.
The dataset contains \num{8861}~manually segmented tree crowns with a \ac{GSD} of \num{1.3} to \SI{3.5}{cm}.
Importantly, the authors note how pre-trained models, such as Detectree2 \cite{ball_accurate_2023} and DeepForest \cite{weinstein_individual_2019}, offer poor performance for mapping dense crowns in their dataset.
While these supervised approaches often yield the best results by relying on annotated data with precise ground truth from expert knowledge, this dependency also limits their real-world applicability, as adapting models to different forest types or shifting identification objectives often requires significant manual photo-interpretation.
Consequently, recent research has sought to mitigate these costs by leveraging alternative data sources. 
Below, we explore the methodologies that alleviate this reliance on manual annotation, namely leveraging unlabelled and synthetic data, targeting data-poor problems such as forestry regeneration.
Accordingly, our work presents an approach that utilizes both a pseudo-label technique and generative AI to address class imbalances and data scarcity in natural regeneration environments, minimizing the need for manual labelling.

\subsection{Automated Labelling with Citizen Science Data}
\label{citizen_science}

Recent research has explored leveraging citizen science data, most notably from the iNaturalist platform \cite{inaturalistcontributors_inaturalist_2026}.
This platform comprises large amounts of images, which are cross-verified by enthusiasts and experts alike, and can be used as training data to learn natural word representations with deep learning models \cite{vanhorn_benchmarking_2021}.
For forest remote sensing applications, \citet{soltani_transfer_2022} first introduced a methodology to leverage an image classifier, trained on iNaturalist data, to generate semantic segmentation masks of high-resolution \ac{UAV} imagery. 
Their study focused on identifying two plant species by training a deep learning model to distinguish the target species from surrounding vegetation.
This classifier was then deployed on orthomosaics with a \ac{GSD} ranging from \num{0.6} to \SI{1.2}{\centi\meter}, via a sliding window approach.
Crucially, this approach generated accurate semantic maps for aerial imagery without requiring any manual pixel-level annotations.
Expanding on this idea, \citet{soltani_simple_2024} applied the same approach to identify ten distinct species using orthomosaics with a finer \ac{GSD} of \SI{0.22}{\centi\meter}.
They further evaluated the efficacy of training a U-Net model \cite{ronneberger_unet_2015} using these automatically generated masks as pseudo-labels for training data, highlighting that spatial resolution is a decisive factor in accurate species identification.

To further explore the effect of \ac{UAV} image spatial resolution, \citet{nasiri_using_2025} utilized a similar workflow to perform semantic segmentation across \num{24} classes, including \numplantclasses plant classes of different taxonomic rank, on even higher resolution imagery. 
They applied an improved technique to generate pseudo-labels for stronger pre-training of a semantic segmentation model. By leveraging over \num{140000}~images of \SI{1}{MP} between \num{0.08} and \SI{0.17}{\centi\meter} \ac{GSD}, they demonstrated that pre-training on pseudo-labels with a Mask2Former \cite{cheng_maskedattention_2022a} architecture can effectively address complex multiclass segmentation. 
The pre-training strategy resulted in significant performance improvements, with an F1 score of \SI{43.74}{\%}, compared to \SI{32.45}{\%} for the baseline supervised approach.
These results demonstrate how leveraging these diverse citizen data repositories can reduce the manual annotation bottleneck found in smaller forest datasets. 
Crucially, their results also indicate that increasing the \ac{GSD} beyond \SI{0.34}{\centi\meter} significantly degrades the performance of the classifier trained on iNaturalist and evaluated on \ac{UAV} images.
Ultimately, while previous methods have established the viability of training perception models without in-domain hand-crafted annotations, they are fundamentally constrained by data availability and diversity.

Our work adopts the pseudo-labelling methodology established by \citet{nasiri_using_2025} to process a significantly expanded corpus of \numaddedunlabelled unlabelled \ac{UAV} images with a spatial resolution of \SI{0.34}{cm} or lower. 
Importantly, relying on automated pseudo-labels from citizen science platforms presents inherent limitations: the data remains noisy, suffers from a significant viewpoint discrepancy compared to drone imagery, and leaves rare plants critically underrepresented.
To actively mitigate these imbalances and improve model generalization, the following section details approaches that leverage synthetic data.
In line with this direction, our work explores the orthogonality of an existing pseudo-label technique with the use of synthetic data.

\subsection{Synthetic Data for Training Neural Networks}

The use of synthetic data as an alternative to real-world observations is a well-established strategy in domains characterized by data scarcity or challenging manual annotation \cite{nikolenko_synthetic_2021}. 
In the context of deep learning for computer vision in forestry, synthetic approaches can be categorized into three different categories: \textit{i)} data augmentation, \textit{ii)} traditional computer graphics pipelines, and \textit{iii)}~\ac{AI}~generation.

Starting with image data augmentations, principally hand-crafted ones, has become a standard technique in deep learning \cite{shorten_survey_2019}, as they are an efficient and scalable way to increase data diversity.
In computer vision, data augmentations are typically performed using random oversampling, where hand-crafted operations, either photometric or geometric, are applied to the image to generate diversity.
Photometric augmentations include modifying brightness, contrast, sharpness, and colours, while geometric augmentations include rotation, translation, and scaling.
More sophisticated techniques blending several training examples, such as CutMix \cite{yun_cutmix_2019}, MixUp \cite{zhang_mixup_2018}, and Cut-Paste and Learn \cite{dwibedi_cut_2017}, have also been proposed. 
Crucially, the choice of augmentations is not trivial \cite{NEURIPS2020_RandAugment}.
Care must also be taken to avoid introducing inconsistencies such as unrepresentative lighting conditions or impossible scene geometries.
It is important to keep in mind that these techniques only modify the 2D pixel-level representation without always preserving the underlying 3D structural coherence of the scene or its semantic content \cite{dvornik_modeling_2018}.
Moreover, augmented data are inherently limited by the diversity of the original data, thus not fully addressing the issue of data scarcity for underrepresented classes.
Despite these constraints, recent work in forestry has adapted the Cut-Paste and Learn augmentation \cite{dwibedi_cut_2017} to generate synthetic training samples from citizen science data. 
\citet{soltani_automated_2025} utilized \ac{Grad-CAM} \cite{selvaraju_gradcam_2017} and the \ac{SAM} \cite{kirillov_segment_2023} to extract plant features from iNaturalist images and composite them onto aerial backgrounds. 
While this bypasses some labelling hurdles, it still relies on the quality of the extraction models and may not capture the complex spatial interactions of a real forest environment.

To circumvent the limitations of existing datasets, researchers have employed traditional computer graphics pipelines to synthesize forest environments, often via 3D game engines.
These approaches rely on explicit geometric modelling and physically based rendering to project 3D structures into 2D images. 
In the context of forestry, \citet{grondin_tree_2023} proposed SynthTree43k, a dataset for tree segmentation and estimation of diameter, felling cut and inclination from ground-level images.
The dataset contains \num{43000} photorealistic images at ground level, which were generated in the Unity game engine with various configurations of trees, landscape, lighting, and weather.
The authors were able to demonstrate precision gains of approximately \SI{6}{\pp} on tree trunk segmentation, a primarily geometric task.
\citet{blaga_forest_2026} introduced the Forest Inspection dataset for \ac{UAV}-based semantic segmentation of forest environments, rendered in the AirSim engine.
This dataset features \num{26319} images taken across a flight sequence with various camera configurations.
The authors report an increase of \SI{1.7}{\pp} in class IoU when pre-training a semantic segmentation network on their synthetic dataset.
However, this data lacks granularity and diversity, as it only presents four plant-based classes in a single forest environment.
The lack of spatial and visual diversity in such simulations stems from the use of assets such as terrain, trees, and foliage, which are frequently generated via procedural algorithms.
Consequently, these simulated environments consistently struggle to bridge the Sim-to-Real gap, lacking the biological variation and photorealism native to regeneration zones.
Because these variations are strictly constrained by explicit geometric rules, they cannot achieve the open-ended, semantically driven diversity that modern prompt-based image generation models can produce.

More recently, advancements in generative AI are enabling new solutions for synthetic data generation \cite{ho_denoising_2020, ramesh_hierarchical_2022}.
Unlike 3D engines, these models do not require explicit geometric modelling; instead, they learn an implicit representation of 3D structure from vast and diverse datasets. 
These methods allow for the creation of new images that maintain the semantic integrity of the target domain, such as specific leaf textures or branching patterns in forestry, while introducing the stochastic variation necessary for deep learning generalization.
For example, \citet{xiang2026doda} developed DODA, an approach which utilizes diffusion models to adapt object detectors to new outdoor agricultural environments.
Instead of relying on manual annotation, these approaches use bounding box layouts to guide the diffusion generation process (i.e., layout-to-image), resulting in synthetic images that are paired with accurate bounding box annotations.
However, bounding boxes do not provide the pixel‑level precision required for detailed species segmentation, particularly in forestry contexts where plants exhibit highly irregular and interwoven spatial structures.
In response to the limitations of coarse annotations, some approaches instead guide the image generation process by providing the model with a semantic segmentation mask.
For instance, \citet{Wang2026SemanticIS} explored the use of a diffusion process to demonstrate the ability to make visually compelling images based on object masks extracted from Cityscapes \cite{Cordts2016Cityscapes} or ADE20K \cite{zhou2018ade20k}.
Similarly, \citet{yang_freemask_2023} demonstrated that diffusion models could generate images from ground-truth semantic masks, yet noted a degradation in quality as scene complexity increased, an issue which would likely be reciprocated for dense natural environments. 

However, all these methods assume the availability of high-quality masks; for orphaned AI projects such as ours, such masks are not available in great quantities.
Consequently, generating photorealistic image and segmentation mask pairs is a highly preferable alternative.
This simultaneous generation was recently explored in medical imaging to tackle similar data scarcity problems.
Early attempts, such as the diffusion-based framework by \citet{bhat_simgen_2025}, struggled with poor generative diversity, exhibiting a strong bias toward the training distribution and a rigid dependence on the initial dataset volume.
To mitigate these issues, \citet{bose_cosimgen_2025} introduced text prompts to inject semantic diversity into the generation process.
Despite improved performance, this method requires training a custom diffusion model from scratch.
As the authors note, such models are inherently data-hungry, creating a circular bottleneck for orphaned domains: a sufficiently large, annotated dataset is required just to train the data generator.
Training such models also requires significant expertise and computing resources, making these approaches unsuitable in practice.

To bypass these limitations, some approaches offer training-free solutions to address long-tailed class imbalances.
For example, MosaicFusion \cite{xie_mosaicfusion_2024} leverages off-the-shelf text-to-image diffusion models to synthesize multi-object scenes from text prompts, extracting corresponding segmentation masks without human intervention by aggregating internal cross-attention maps.
To increase production throughput, the authors propose generating spatial mosaic images, paired with the semantic masks.
However, it has only been evaluated on sparse scenes, typically featuring a single object per sub-image and a maximum of four sub-images per mosaic.
\textit{Gen-n-Val} \cite{Huang2025GennValAI} further built on MosaicFusion by first improving the prompting and then by obtaining better instance segmentation of the generated objects via a \ac{VLM}.
Yet, these approaches still have limitations for the generation of dense semantic segmentation images and masks.
For example, our real annotated images have an average of 14.9 distinct instances, which would be hard to replicate with this approach.
In our context, we define an instance as a spatially isolated cluster of connected pixels belonging to the same semantic class.
For semantic segmentation, \citet{wu2023diffumask} developed DiffuMask, which extracts pixel-level masks directly from the model's internal cross-attention maps during the image generation process.
The authors use a combination of CLIP \cite{radford_learning_2021} and Stable Diffusion \cite{rombach_highresolution_2022} to generate images from text prompts.
However, the low resolution of the attention maps results in coarse masks that likely would not capture the fine-grained details of forest \ac{UAV} images.
Moreover, the authors only evaluate image generation for a few well-defined and distinct objects.

More recently, large-scale vision-language models have emerged as a complementary paradigm for image generation.
Models such as DALL-E~2 \cite{ramesh_hierarchical_2022} and the more recent Nano Banana Pro \cite{google_gemini_2026} are trained on millions or even billions of images and an enormous amount of text, enabling the generation of photorealistic images across a wide range of classes and contexts.
A key characteristic of these models is their ability to be guided by natural-language and image prompts, allowing users to specify scene content, structure, and classes.
Their broad visual and semantic coverage, combined with prompt-based control, suggests an increased capacity to represent the variability and complexity of natural forest environments without requiring task-specific training.
Reflecting these capabilities, the use of such a large-scale vision-language model has been explored in ecological remote sensing.
Notably, \citet{durand_lacking_2026} demonstrated the use of DALL-E~2 to generate synthetic aerial imagery of muskox, addressing the limited availability of observations for this species.
Their results highlight the potential of large multimodal models to support data generation in domains where data acquisition is constrained.
Crucially, the authors relied on manual annotation of the synthetic images, which is a severe bottleneck for tasks such as ours, where expert knowledge is often required.

Building on these developments, our work investigates the use of the Nano Banana Pro \cite{google_gemini_2026} model for the complex task of fine-grained forest monitoring, diverging from traditional supervised trajectory methods by prioritizing data efficiency over extensive manual annotation.
To this end, we contribute an AI-generated dataset, \GenAiDataset, comprising \numregen synthetic images and corresponding segmentation masks.
We present a comparison of our approach for joint image and segmentation mask generation against other techniques in \autoref{tab:synthetic_methods_table}.
Notably, prior works suffer from distinct trade-offs: they either lack photorealism, rely on training specialist models, are limited to simple contexts, or do not include joint generation of segmentation masks.
In contrast, our approach simultaneously and elegantly satisfies all four criteria.
The \GenAiDataset dataset, generated through our proposed approach, contains high-quality, dense, and complex images of forest regeneration scenes.
We assess the performance of semantic segmentation deep learning methods when augmented with our synthetic data and conduct a zero-shot evaluation to determine the inherent capacity of general-purpose generative models to produce accurate semantic masks.
Most importantly, models such as Nano Banana Pro have reached a level of scene understanding which could unlock a new paradigm for training deep learning models.
These models require little expertise and are trained on so much data that they exhibit excellent zero-shot scene understanding across many different contexts \cite{gabeur_image_2026}.

\begin{table}[ht]
\centering
\caption{Comparison of approaches for synthetic data generation.}
\label{tab:synthetic_methods_table}
\footnotesize
\setlength{\tabcolsep}{6pt}
\begin{tabular}{l c c c c}
    \toprule
    \textbf{Approach} & \textbf{Photorealistic} & \textbf{Zero-Shot} & \textbf{Complex} & \textbf{Mask} \\
                      & \textbf{Images}         &                    & \textbf{Scenes}  & \textbf{Generation} \\
    \midrule
    Data Augmentation & \multirow{2}{*}{--} & \multirow{2}{*}{$\checkmark$} & \multirow{2}{*}{--} & \multirow{2}{*}{$\checkmark$} \\
    \cite{soltani_automated_2025} & & & & \\
    \midrule
    Traditional Computer Graphics & \multirow{2}{*}{--} & \multirow{2}{*}{$\checkmark$} & \multirow{2}{*}{$\checkmark$} & \multirow{2}{*}{$\checkmark$} \\
    \cite{grondin_tree_2023, blaga_forest_2026} & & & & \\
    \midrule
    Generative AI: & & & & \\
    \quad Mask-to-Image & \multirow{2}{*}{$\checkmark$} & \multirow{2}{*}{--} & \multirow{2}{*}{--} & \multirow{2}{*}{--} \\
    \quad \cite{xiang2026doda, Wang2026SemanticIS, yang_freemask_2023} & & & & \\
    \addlinespace[0.3em]
    \quad Cojoint Diffusion & \multirow{2}{*}{$\checkmark$} & \multirow{2}{*}{--} & \multirow{2}{*}{--} & \multirow{2}{*}{$\checkmark$} \\
    \quad \cite{bhat_simgen_2025, bose_cosimgen_2025} & & & & \\
    \addlinespace[0.3em]
    \quad Attention Extraction & \multirow{2}{*}{$\checkmark$} & \multirow{2}{*}{$\checkmark$} & \multirow{2}{*}{--} & \multirow{2}{*}{$\checkmark$} \\
    \quad \cite{xie_mosaicfusion_2024, Huang2025GennValAI, wu2023diffumask} & & & & \\
    \addlinespace[0.3em]
    \quad Image Generator & \multirow{2}{*}{$\checkmark$} & \multirow{2}{*}{$\checkmark$} & \multirow{2}{*}{$\checkmark$} & \multirow{2}{*}{--} \\
    \quad \cite{durand_lacking_2026} & & & & \\
    \midrule
    \textbf{Ours} & $\mathbf{\checkmark}$ & $\mathbf{\checkmark}$ & $\mathbf{\checkmark}$ & $\mathbf{\checkmark}$ \\
    \bottomrule
\end{tabular}
\end{table}

\section{Methodology}

In this section, we first introduce the datasets utilized in our study, detailing our extension of the \nasiridataset dataset, both its hand-labelled and unlabelled subsets.
We also detail the creation of our novel \GenAiDataset dataset that actively addresses class imbalance and labelled data scarcity.
To the best of our knowledge, this is the first dataset exploring the use of a large-scale vision model to produce complex synthetic images paired with precise semantic masks.
Next, we follow up with an overview of the neural network architectures selected for evaluation.
Finally, we outline our training and evaluation methodology, which emphasizes a rigorous assessment of model generalization toward unseen regions.

\subsection{Datasets}

We propose two datasets for forest regeneration mapping.
First, our \ourdataset dataset is an extension of \nasiridataset \cite{nasiri_using_2025}.
It contains a total of \numunlabelled unlabelled and \numlabelled hand-labelled \ac{UAV} images collected in post-disturbance forest regeneration zones.
Importantly, this dataset provides an ideal proving ground for generative AI by deviating from standard semantic segmentation benchmarks: it encompasses \numsitestotal~geographically diverse areas, \numclasses~non-standard classes exhibiting class imbalance, all collected in open-world conditions with plenty of out-of-distribution observations.
Second, we propose the \GenAiDataset dataset containing \numregen AI-generated images and pixel-aligned semantic masks.
Crucially, we leverage this dataset to address the issues of labelled data scarcity and class imbalance.

\subsubsection{\ourdataset{} -- Hand-labelled Subset}

For our hand-labelled data subset, we propose an extended version of the hand-labelled subset in \nasiridataset \cite{nasiri_using_2025}.
To analyze forest regeneration, we group the data into \numclasses classes, consisting of \numplantclasses plant classes and \numnonplantclasses non-plant classes.
The selection of these categories is guided by both their abundance in the surveyed regions and their ecological roles, including economically important tree species alongside competing flora and surface-level obstructions \cite{ministeredesressourcesnaturellesetdelafaune_sustainable_2026}.
Because \ac{UAV} imagery does not allow consistent species-level classification for smaller plants, categories vary in taxonomic specificity: larger plants can be distinguished at the species level, while lower-growing ones are represented at broader levels such as the genus.
The full classification hierarchy is described below:
\begin{itemize}[leftmargin=*,labelsep=1em,topsep=5pt]
  \item \textbf{Division:} Moss (\textit{Bryophyta}).
\vspace{-4pt}
  \item \textbf{Class:} Fern (\textit{Polypodiopsida}).
\vspace{-4pt}
  \item \textbf{Family:} Sedge (\textit{Cyperaceae}).
\vspace{-4pt}
  \item \textbf{Genus:} Fir (\textit{Abies}), Serviceberry (\textit{Amelanchier}), Willowherb (\textit{Epilobium}), Spruce (\textit{Picea}), Pine (\textit{Pinus}).
\vspace{-4pt}
  \item \textbf{Species:} Red Maple (\textit{Acer rubrum} L.), Mountain Maple (\textit{Acer spicatum} Lam.), Yellow Birch (\textit{Betula alleghaniensis} Britt.), Paper Birch (\textit{Betula papyrifera} Marsh.),  Sheep Laurel (\textit{Kalmia angustifolia} L.), Trembling Aspen (\textit{Populus tremuloides} Michx.), Fire Cherry (\textit{Prunus pensylvanica} L.f.), Bog Labrador Tea (\textit{Rhododendron groenlandicum} (Oeder) Kron \& Judd), Red Raspberry (\textit{Rubus idaeus} L.), American Mountain-Ash (\textit{Sorbus americana} Marsh.), Canadian Yew (\textit{Taxus canadensis} Marsh.), Lowbush Blueberry (\textit{Vaccinium angustifolium} Aiton).
\vspace{-4pt}
  \item \textbf{Non plant classes:} Wood, Boulder, Other.
\end{itemize}

Compared to \nasiridataset, the most notable changes were to remove an error-prone class, Dead Tree, and strengthen the annotations to ensure consistency across images.
Indeed, the Dead Tree class was ambiguous during the annotation process, resulting in neural network models showing confusion between the Dead Tree and Wood classes \cite{nasiri_using_2025}. 
Given the difficulty of distinguishing these classes from \ac{UAV} images in forest regeneration zones, the Dead Tree class was simply merged into the Wood class.
Furthermore, four images, which had a high representation of the \textit{Viburnum} genus and were thus heavily biased towards the Other class, were removed.
Finally, we annotated \numaddedlabelled images from previously unlabelled images in \nasiridataset, which contain rarer classes such as Canadian Yew and Pine.
Each image is a cropped region from our full-resolution UAV images, which range from \num{8064}~$\times$~\num{6048} to \num{4000}~$\times$~\num{2250} pixels, down to \num{1024}~$\times$~\num{1024} pixels.
In total, our extended dataset contains \numlabelled hand-labelled images.
Despite the increased amount of hand-labelled data, class imbalance remains a significant issue.
As such, this dataset is well-suited to benchmark our proposed techniques to mitigate class imbalance.

\subsubsection{\ourdataset{} -- Unlabelled Subset}

\begin{figure}[!t]
    \centering{\includegraphics[width=1.0\textwidth]{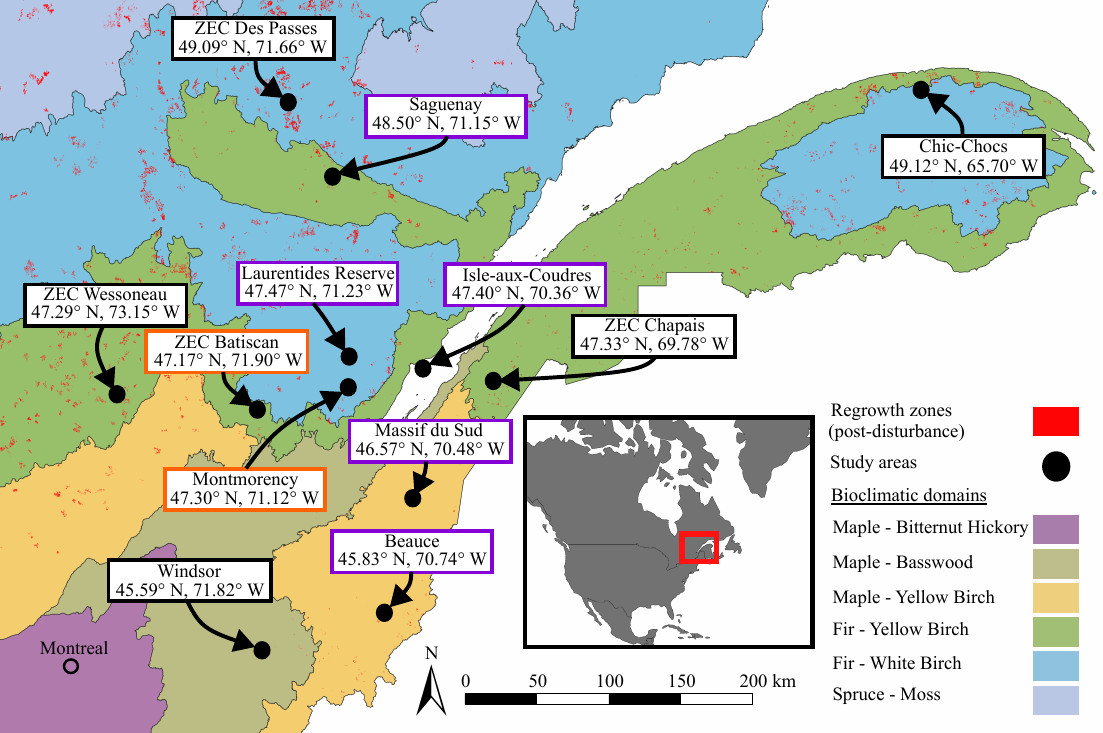}}
    \caption{Map of the different data collection locations present in \ourdataset. Sites visited for \nasiridataset are shown in a black box \textcolor[RGB]{0, 0, 0}{\rule{10pt}{6pt}}, sites revisited for \ourdataset are shown in a orange box \textcolor[RGB]{255, 97, 5}{\rule{10pt}{6pt}}, and new sites are shown in a purple box \textcolor[RGB]{133, 0, 214}{\rule{10pt}{6pt}}.
    Data collection locations were in zones logged between 2018 and 2023, with vegetation growing for two to six years, and were visited during the summer in 2023, 2024, or 2025.}
    \label{fig:data_map}
\end{figure}

The unlabelled subset used in this study expands upon the \nasiridataset dataset \cite{nasiri_using_2025} by incorporating \numaddedunlabelled new images, bringing the total number of images to \numunlabelled.
To optimize data collection, we increased the \ac{UAV} flight altitude while strictly adhering to critical resolution thresholds.
\citet{nasiri_using_2025} demonstrated that model performance degrades significantly at a \ac{GSD} coarser than \SI{0.34}{cm}, with little gains at finer resolutions.
Consequently, our dataset extension of \numaddedunlabelled images was captured at altitudes calibrated to maintain a maximum \ac{GSD} of \SI{0.34}{cm}, thereby maximizing spatial coverage without compromising identification accuracy.
To capture a broader range of phenotypic and structural variation within our target plant classes, these additional images were collected in diverse geographic sites, including a new bioclimatic domain, Maple - Yellow Birch.
Bioclimatic domains represent distinct ecoregions defined by Quebec's Ministry of Natural Resources and Forests based on their unique climate, vegetation, and characteristics \cite{ministere_des_ressources_naturelles_et_des_forets_zones_2022}.
\autoref{fig:data_map} shows a map of the different data collection sites across the province of Quebec, both for \nasiridataset and \ourdataset.
Importantly, the full dataset contains images from \numsitestotal different locations and \numbioclimdomains bioclimatic domains.
Moreover, we diversified the image sensors used to collect the added unlabelled images. 
The resulting unlabelled subset features \numunlabelled images acquired across multiple field deployments using a variety of sensors: a \SI{12}{MP} DJI Mini 2 camera, a \SI{20}{MP} DJI Mavic 3 Enterprise camera, a \SI{12}{MP} DJI Mavic 3 Enterprise 7$\times$~zoom camera, and a \SI{48}{MP} DJI Mini 4 Pro camera.
We note that the \numunlabelled full-size \ac{UAV} images are equivalent to over \SI{550000}{MP}, approximately \num{2750}$\times$ larger than our hand-labelled subset.
Using the pseudo-annotation technique based on a sliding window classification method described in \citet{nasiri_using_2025} and detailed later in \refSection{sec:pseudolabelgeneration}, this dataset will serve as an alternate source of training labels for our experiments.
In particular, we will be able to empirically evaluate the orthogonality of AI-generated images with respect to pseudo-labelled real images.

\subsubsection{\GenAiDatasetName}
\label{sec:GenAIgeneration}

\begin{figure}[!h]
    \centering
    \includegraphics[width=1.0\textwidth]{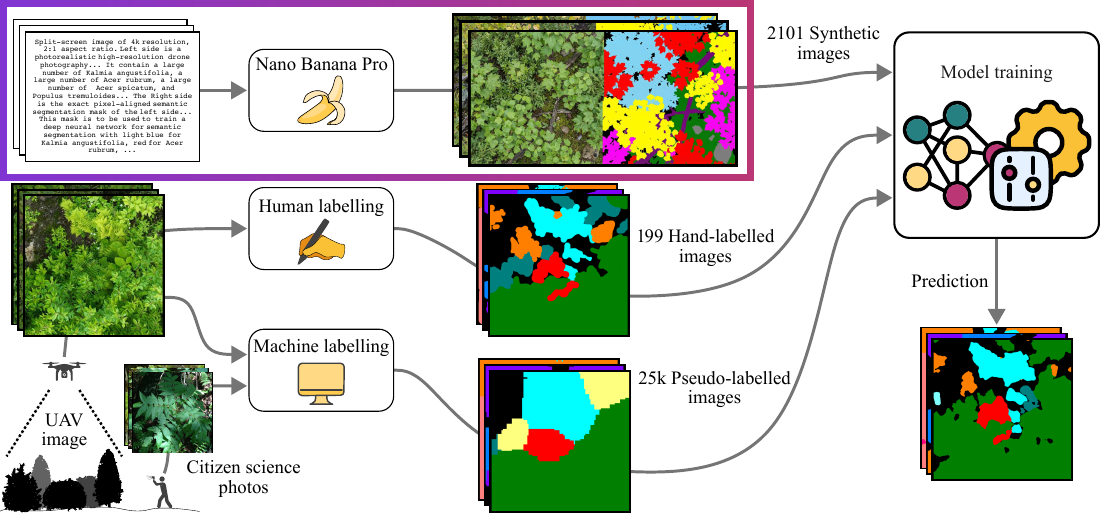}
    \caption{Overview of our model training pipeline, combining \numlabelled hand-labelled and \numunlabelled pseudo-labelled real images, and \numregen synthetic image and label pairs.
    The pseudo-labelled real images are equivalent to over \num{550000} image crops of 1 MP.
    The top left (purple outline) represents our primary contribution, the \GenAiDataset dataset, which serves to demonstrate that image generators can now produce photorealistic images and corresponding annotations as viable training data for semantic segmentation. 
    For this, we leverage the Nano Banana Pro model to produce synthetic, top-view images and masks of forest regeneration. 
    This component is key in addressing the scarcity of labelled data and class imbalance.
    }
    \label{fig:overview}
\end{figure}

Despite extending our unlabelled and hand-labelled subsets, high class imbalance and under-representation remain.
Furthermore, deep learning models require large amounts of diverse data to properly adapt to new environments and contexts \cite{beery_recognition_2018}, but \textit{in~situ} data collection can still be costly.
To address this, we leverage the advanced semantic capabilities of image generators, which can accurately generate complex images from natural language prompts.
We introduce the \GenAiDataset dataset, composed of \numregen synthetic images \textit{and} labels generated with the Nano Banana Pro model\footnote{\texttt{gemini-3-pro-image-preview}, available from November 20, 2025 to July 17, 2026.} from \citet{google_gemini_2026}.
Synthetic images and their respective multi-class segmentation masks were generated \textit{in a single inference cycle}, with prompts instructing the model to assign distinct colour identifiers to each class present in the scene for mask generation, similarly to \citet{gabeur_image_2026}.
These generated images and masks can then be used to train a semantic segmentation neural network alongside our hand-labelled and pseudo-labelled \ac{UAV} images.
An overview of the data sources and labelling techniques used to train our models can be found in \autoref{fig:overview}.
A detailed analysis of the differences between labelled and synthetic images, along with variations in mask complexity, is presented in \autoref{sec:genai_dataset_analysis}.

\begin{figure}[!h]
    \centering
    \includegraphics[width=1.0\linewidth]{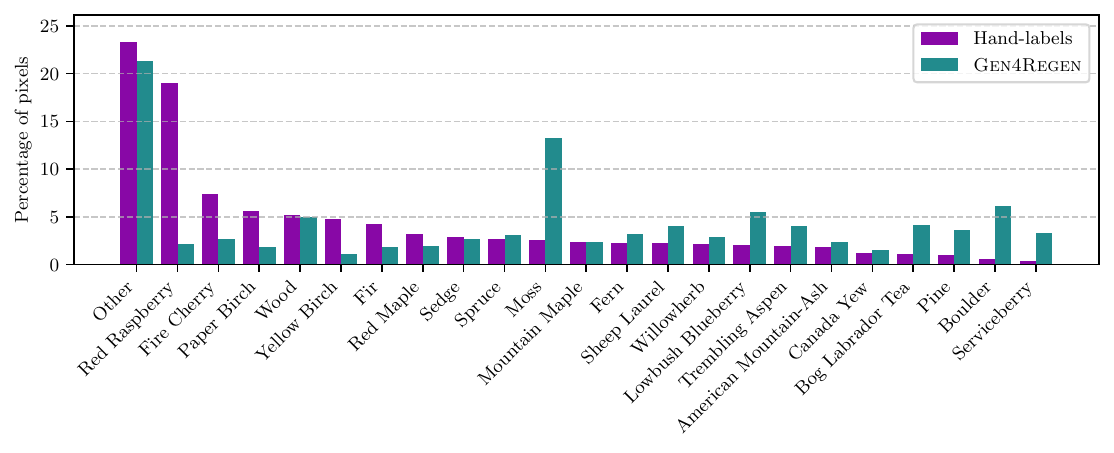}
    \caption{Pixel distributions of our hand-labelled subset in \ourdataset and our \GenAiDataset dataset.
    The hand-labelled subset of real images exhibits a strong class imbalance, despite our efforts to find appropriate images to annotate in the \num{25000}+ captured \ac{UAV} images.
    Meanwhile, \GenAiDataset directly addresses this issue by focusing on underrepresented classes.}
    \label{fig:class_distribution}
\end{figure}

Our prompts were carefully designed to increase diversity and fidelity alongside our existing data, most notably across underrepresented classes.
The class distribution of our hand-labelled subset in \ourdataset and our \GenAiDataset dataset is depicted in \autoref{fig:class_distribution}.
To ensure dataset diversity, our prompting mechanism utilized stochastic attribute sampling, where environmental variables were randomized, including seasons, moisture levels, and lighting conditions.
We favoured this approach instead of using an \ac{LLM} to create the prompts \cite{Wen2023FoundationPoseU6, Huang2025GennValAI}, to ensure that the prompts were well-aligned with our image domain.
We explicitly enforced a top-view perspective and low \ac{GSD} to simulate low-altitude \ac{UAV} captures, as in our \ourdataset dataset.
For more information about the prompts used, \autoref{sec:PromptGenerationTechnique} shows different prompt examples alongside the resulting generations.

Once the generation was complete, every generated image-mask pair underwent a rapid visual inspection to filter out hallucinations, missing masks, spatial misalignment, mask colours appearing in the image, incorrect semantic assignments, or incorrect viewpoint.
Several examples of failure cases are shown in \autoref{sec:generation_failure_cases}.
This quality assurance step required only 3 to 30 seconds per sample, depending on the subtlety of the defect, which is about two orders of magnitude quicker than manual annotation at around 20 minutes.
Consequently, an operator without specialized domain knowledge can generate and curate hundreds of fully annotated images per hour.
This approach is in stark contrast to manual semantic segmentation annotation, which typically yields fewer than three images per hour while strictly demanding expert-level forestry proficiency.
In addition to removing generation failure cases, correct generations were automatically processed to remove generation artifacts, including digital watermarks.
The generated masks were finally post-processed to assign the colours to the corresponding class, before adding them to the \GenAiDataset dataset.

Overall, the generative model demonstrated a high success rate, with over \SI{85}{\%} of the generated outputs meeting the quality requirements for inclusion in our \GenAiDataset dataset. 
Empirical observations revealed specific limitations regarding prompt complexity.
Notably, requesting more than five distinct plants simultaneously led to a high failure rate, with a majority of generated images containing mistakes. 
Consequently, the maximum number of co-occurring classes per image was constrained to ensure label fidelity. 
Fortunately, the average number of classes in our \ourdataset labelled dataset is 3.8, well-aligned with this limitation.
To streamline the data curation process, images were generated in batches of 50 to 100 using a fixed subset of classes.
This strategy mitigated challenges associated with the model's tendency to output approximate RGB values for the segmentation masks, thereby simplifying the necessary post-extraction mask label verification.
Furthermore, the model exhibited sensitivity to label aggregation when trying to explicitly merge subspecies at prompt time.
For instance, prompting the generation of three distinct blueberry species (\textit{Vaccinium angustifolium}, \textit{Vaccinium myrtilloides}, and \textit{Vaccinium boreale}) under a single, unified mask colour resulted in frequent mask generation errors.
To circumvent this limitation, the model was instructed to assign distinct colour gradients (e.g., varying shades of blue) to each subspecies, which were subsequently merged into a single semantic class during post-processing.

\subsection{Model Architectures}

To evaluate the impact of neural network choice on forest regeneration monitoring, we evaluate three state-of-the-art architectures: Mask2Former \cite{cheng_maskedattention_2022a}, DINOv2 \cite{oquab_dinov2_2024}, and DINOv3 \cite{simeoni_dinov3_2025}.
By selecting these architectures, we aim to assess both task-specific architectural innovations and the generalization capability of large-scale foundation backbones.
We adopt Mask2Former, as it employs a mask classification paradigm that is highly effective for delineating complex boundaries in high-resolution imagery. 
In contrast to traditional pixel-wise classification, this architecture utilizes a Transformer-based decoder with masked attention to focus on localized features, making it particularly suitable for identifying distinct plants in forest regeneration zones. 
For our experiments, we utilize the Large variant (i.e., \num{215} million parameters) of this architecture, with a hierarchical Swin Transformer backbone \cite{liu_swin_2021}.
As a complementary approach, we use DINOv2 and DINOv3 as robust foundation backbones.
Both backbones are \acp{ViT} pre-trained via self-supervised learning on massive curated datasets, resulting in versatile feature representations that generalize well to specialized domains such as forestry.
This pre-training allows the backbones to distinguish between visually similar plants even when localized training data is limited.
More specifically, DINOv3 is an incremental step over DINOv2, with improvements to pre-training and architecture.
Given the relative simplicity of upgrading backbones over time, we chose to experiment with both to establish if these improvements would translate to free performance gains on our task.
To leverage the DINO backbones, we use a \ac{ViT}-based dense prediction head \cite{ranftl_vision_2021} for semantic segmentation prediction.
Similar to the Mask2Former model, we employ the Large variants of the DINO backbones (i.e., \num{334} million parameters), which provide a strong balance between computational efficiency and the depth of learned features required for ecological monitoring.
In addition, following the semantic segmentation evaluation protocols for DINO backbones established in \cite{oquab_dinov2_2024, simeoni_dinov3_2025}, we freeze the backbone weights. 
This technique prevents drastic changes to the strong features already learned by the model while significantly reducing training time, with a smaller number of trainable parameters (i.e., \num{31} million).

\subsection{Pseudo-label Generation on Unlabelled Images}
\label{sec:pseudolabelgeneration}

We seek to understand the orthogonality of different data sources for model training.
By implementing the pseudo-labelling methodology of \citet{nasiri_using_2025} on our unlabelled \ac{UAV} imagery, we designed our evaluation to quantify whether combining synthetic and pseudo-labelled data produces genuinely additive performance gains.
To create the pseudo-labels, we first train a DINOv3 \cite{simeoni_dinov3_2025} classification model on a curated iNaturalist dataset, which is almost identical to the one used in \cite{nasiri_using_2025}; our only change is the removal of the Dead Tree class.
Our iNaturalist subset used to train this classifier thus contains \numinaturalist images taken by citizens, split over our \numclasses classes.
Note that it only contains \texttt{Research Grade} images, which are subsequently filtered using a neural network classifier to remove low-quality images that either have limited visibility or feature outliers.
The \texttt{Research Grade} of images corresponds to a subset of iNaturalist observations where the species identification has been agreed upon by the community.
We then applied our DINOv3 classifier on our \numunlabelled unlabelled \ac{UAV} images via a sliding window approach to obtain pseudo-labels.
However, instead of using \SI{1}{MP} crops of the images for pseudo-label generation as in \citet{nasiri_using_2025}, the pseudo-labels were created on the full \ac{UAV} images.
In fact, full-size images contain a wider view and hence more information, which could be beneficial for model training and inference.
The resulting pseudo-labelled dataset contains \numunlabelled full-size \ac{UAV} images ranging from \num{12} to \SI{48}{MP}, which is equivalent to over \SI{550000} image crops of \SI{1}{MP}.

\subsection{Model Training and Evaluation}

When using training data from different sources, such as AI-generated and real images, it is important to mitigate domain discrepancies during training to avoid unforeseen biases which could be detrimental to generalization. 
In our situation, the mixing ratio between real and AI-generated images is a key hyperparameter.
Through hyperparameter search, we identified a 40:60 synthetic:real image ratio as a good balance between both data sources.
Lower ratios, such as 25:75, were tested, but limited the contribution of synthetic data, while higher ratios, such as 50:50, were detrimental to model performance.
Consequently, the 40:60 ratio was used in all experiments.
In addition, we tested three different batching strategies to combine labelled and synthetic data.
First, we tested a homogeneous batch strategy, in which each batch contains either labelled or synthetic data exclusively.
The image ratio was then imposed on the batches themselves.
Second, we tested a balanced heterogeneous batch strategy, which balances the contents of each batch with the image ratio.
Third, we tested a weighted sampling strategy, which randomly samples real or synthetic images for each batch with a probability determined by the image ratio.
Following a hyperparameter search, we found that the weighted sampling strategy yielded slightly stronger performance, although the difference was not statistically significant.
This sampling strategy was still used for all experiments as, in contrast to the homogeneous and balanced heterogeneous approaches, it maintained the desired synthetic-to-real ratio while introducing greater batch-to-batch diversity during training.
In addition to the hand-labelled and \GenAiDataset data sources, pseudo-labels were also used as pre-training data, following the methodology of \citet{nasiri_using_2025}.

To conduct the experiments with Mask2Former, we use the Hugging Face implementation from the Transformers library \cite{wolf-etal-2020-transformers}, and the model weights are initialized from pre-training with the ADE20K dataset \cite{zhou2018ade20k}.
The model uses a learning rate of $10^{-4}$ with a backbone factor of 0.1 and a weight decay of 0.05.
Training is performed with a batch size of 16.
The loss function is a combination of class, mask, and dice weights, set at 2.0, 5.0, and 5.0, respectively. 
For the DINO models, we use the implementation from \citet{yang_unimatch_2025} with the standard weights \cite{oquab_dinov2_2024, simeoni_dinov3_2025}.
The model uses a learning rate of $2 \times 10^{-4}$.
A per-pixel cross-entropy loss is used with a weight decay of 0.01.
The batch size for training is also 16.
Training was done on a compute server with eight NVIDIA A100 GPUs, each with \SI{80}{GB} of VRAM.

To capitalize on the expanded labelled dataset and provide stronger statistical validation than the single test split used in \citet{nasiri_using_2025}, we implemented a rigorous five-fold cross-validation strategy. 
Folds were constructed with data from geographically distinct collection areas separated by at least \SI{20}{\kilo\meter}, rather than randomly distributing the data across folds, while striving for comparable class representations, despite strong imbalance.
This spatial segregation serves multiple crucial research objectives.
First, it explicitly prevents geographical data leakage, ensuring that the test sets do not share the same species compositions, environments, \ac{UAV} flight trajectories, weather, or illumination conditions as the training data, thereby severely stressing the network's capacity to generalize.
Likewise, when pre-training on the pseudo-labelled dataset, we ensure that no pseudo-labels used for training are geographically close to the validation or test set. 
This approach is a departure from \citet{nasiri_using_2025}, who pre-trained on the entirety of the pseudo-labels.
Second, evaluating our models across five distinct folds significantly enhances the statistical reliability of our findings, confirming that performance gains are not artifacts of a specific data split. 
Third, by grouping images regionally, this spatial splitting naturally induces a realistic degree of class imbalance, accurately mirroring the challenging conditions for practitioners to train networks for real-world, data-scarce applications.
We compute final performance metrics by pooling predictions from all folds across the full dataset. 
To account for the high variance inherent in training deep learning models on small datasets, we perform eight independent runs with distinct random seeds.
This allows for a more stable statistical evaluation of model performance than a single five-fold pass.
We also present the standard deviation to quantify the variance.

To evaluate model performance under severe class imbalance, we report three distinct segmentation metrics: macro-averaged F1 score, \ac{mIoU}, and global pixel accuracy.
The macro-averaged F1 score is computed by averaging the per-class F1 scores across all classes: $$F1_{macro} = \frac{1}{C} \sum_{c=1}^{C} \frac{2TP_c}{2TP_c + FP_c + FN_c},$$ where $TP_c$, $FP_c$, and $FN_c$ denote the number of true positive, false positive, and false negative pixels for the class $c$ respectively.
True positives correspond to pixels correctly classified as class $c$; false positives are pixels incorrectly predicted as class $c$ but belonging to other classes (commission errors); and false negatives are pixels belonging to class $c$ that are incorrectly assigned to another class (omission errors).
Similarly, \ac{mIoU} averages the intersection over the union of the ground truth and predicted mask for each class $c$.
Mathematically, $$mIoU = \frac{1}{C} \sum_{c=1}^{C} \frac{TP_c}{TP_c + FP_c + FN_c}.$$
The global pixel accuracy is computed as $$Pixel~accuracy = \frac{Number~of~correctly~classified~pixels}{Total~number~of~pixels}.$$
We prioritize the macro F1 score and \ac{mIoU} as our primary evaluation metrics.
By treating every class with equal weight regardless of its pixel frequency, these macro-averaged metrics prevent the performance on dominant classes from masking classification results on rare classes.
We also report global pixel accuracy for completeness, but we treat it strictly as a secondary baseline metric due to its vulnerability to class imbalance.

\section{Results}

To evaluate the use of AI-generated data for semantic segmentation in the context of post-disturbance forest regeneration monitoring, we first assess the performance of the Mask2Former, DINOv2, and DINOv3 models with multiple combinations of training data, namely \textit{i)} hand-labelled, \textit{ii)} pseudo-labelled, and \textit{iii)} AI-generated.
Then, we follow up with additional experiments on the best-performing model.
We assess the segmentation performance when labelled data is unavailable to show how one can bootstrap semantic segmentation projects without the need for real annotated data.
Next, we evaluate the zero-shot semantic segmentation capabilities of the Nano Banana Pro model on our \ac{UAV} images.
We continue with an ablation study showing the performance of the model when reducing the number of \GenAiDataset images during training.
Finally, we present some qualitative results for semantic segmentation when training on different data sources.

\subsection{Evaluating the Contribution of Each Data Source to Network Performance}

We conducted a series of experiments to isolate and quantify the individual contributions of each of the three datasets to the final model performance.
Specifically, we established the manually labelled data as our baseline and systematically evaluated the impact of incorporating the remaining two sources: i) introducing a pre-training phase with our vast pseudo-labelled corpus, equivalent to over 550k images of \SI{1}{MP}, and ii) augmenting the primary training set with \GenAiDataset.
The impact of these different data combinations on semantic segmentation performance is summarized in \autoref{tab:results}.
For each combination of training data, we experimented with the Mask2Former, DINOv2, and DINOv3 models.
Results show how, independently of the model used, pseudo-labels and \GenAiDataset both help model generalization and performance. 
Interestingly, both data sources are orthogonal when combined, showing over \SI{15}{\pp} gain compared to training the model only on hand-labelled data.
For instance, Mask2Former gains \SI{5.60}{\pp} with \GenAiDataset and \SI{10.56}{\pp} with pseudo-labels.
Coincidentally, the sum of these gains is very close to the \SI{17.33}{\pp} achieved by training with both modalities, demonstrating the aforementioned orthogonality.
Likewise, DINOv2 gains \SI{8.86}{\pp} with \GenAiDataset, \SI{8.40}{\pp} with pseudo-labels, while training with both leads to a gain of \SI{13.83}{\pp}.
For DINOv3, the gains of adding pseudo-labels and \GenAiDataset are not as orthogonal, but still show improved performance when combined, with a gain of \SI{15.17}{\pp} compared to training only on the labelled dataset.
These results also demonstrate that Mask2Former benefits more from the addition of pseudo-labels than \GenAiDataset.
We argue that Mask2Former’s mask classification architecture explicitly prioritizes mask quality, making it more receptive to the strong class-level supervision provided by pseudo-labels.
In contrast, when accounting for standard deviation, the DINO models seem to benefit equally from the fine segmentation masks of the \GenAiDataset dataset and the strong class information found in the pseudo-labels.
Crucially, DINOv3 obtains significant performance gains over DINOv2, between \num{5} and \SI{7}{\pp} across all metrics.
As a reminder, DINOv3 mainly consists of incremental improvements for pre-training and backbone architecture.
Given the relative simplicity of swapping a model backbone, it is promising to see that this can lead to free performance gains over time, as similar foundational backbones are regularly being developed.
Finally, we note that DINOv3 achieves the best results, with \SI{60.75}{\%} of F1 score, \SI{46.42}{\%} of \ac{mIoU}, and \SI{68.20}{\%} of pixel accuracy.
These results highlight the complexity of semantic segmentation of plants in forest regeneration zones, as our best results, which leverage our hand-labelled, pseudo-labelled, and AI-generated data, leave ample room for improvement.

\begin{table}[!h]
    \centering
    \caption{Ablation study using different models and data sources for semantic segmentation. 
    The metrics $\pm$ standard deviation are presented.
    For each model, the best results are highlighted in \textbf{bold}, and second-best results are \underline{underlined}.
    \label{tab:results}}
    {\begin{tabular}{lcccccc} \toprule
    Model & Hand-labels & Pseudo-labels & \GenAiDataset & F1 score & mIoU & Pixel \\
    & (\numlabelled images) & (\numunlabelled images) & (\numregen images) &  & & Accuracy \\
    \midrule
    
    Mask2Former & \checkmark & -- & -- & 42.04$_{\pm 0.96}$ & 31.43$_{\pm 0.72}$ & 59.42$_{\pm 0.92}$ \\
    & \checkmark & \checkmark & -- & \underline{52.60}$_{\pm 0.46}$ & \underline{39.91}$_{\pm 0.37}$ & \underline{66.73}$_{\pm 0.22}$ \\
    & \checkmark & -- & \checkmark & 47.64$_{\pm 2.35}$ & 35.62$_{\pm 1.75}$ & 61.74$_{\pm 0.87}$ \\
    & \checkmark & \checkmark & \checkmark & \textbf{59.37}$_{\pm 0.96}$ & \textbf{45.13}$_{\pm 0.97}$ & \textbf{68.74}$_{\pm 0.41}$ \\
    \midrule
    DINOv2 & \checkmark & -- & -- & 39.46$_{\pm 0.86}$ & 28.27$_{\pm 0.72}$ & 56.55$_{\pm 0.96}$ \\
    & \checkmark & \checkmark & -- & 47.86$_{\pm 0.89}$ & \underline{35.62}$_{\pm 0.68}$ & \underline{62.48}$_{\pm 0.47}$ \\
    & \checkmark & -- & \checkmark & \underline{48.32}$_{\pm 1.69}$ & 35.42$_{\pm 1.21}$ & 60.23$_{\pm 0.48}$ \\
    & \checkmark & \checkmark & \checkmark & \textbf{53.29}$_{\pm 0.93}$ & \textbf{39.25}$_{\pm 0.84}$ & \textbf{63.16}$_{\pm 0.82}$ \\
    \midrule
    DINOv3 & \checkmark & -- & -- & 45.58$_{\pm 1.38}$ & 34.04$_{\pm 1.14}$ & 61.84$_{\pm 1.12}$ \\
    & \checkmark & \checkmark & -- & 55.56$_{\pm 0.84}$ & 41.71$_{\pm 0.64}$ & \underline{66.86}$_{\pm 0.35}$ \\
    & \checkmark & -- & \checkmark & \underline{57.35}$_{\pm 1.17}$ & \underline{43.33}$_{\pm 1.00}$ & 65.92$_{\pm 0.58}$ \\
    & \checkmark & \checkmark & \checkmark & \textbf{60.75}$_{\pm 0.69}$ & \textbf{46.42}$_{\pm 0.78}$ & \textbf{68.20}$_{\pm 0.51}$ \\
    \midrule
    \end{tabular}}{}
\end{table}

\begin{figure}[h!]
    \centering{\includegraphics[width=1.0\textwidth]{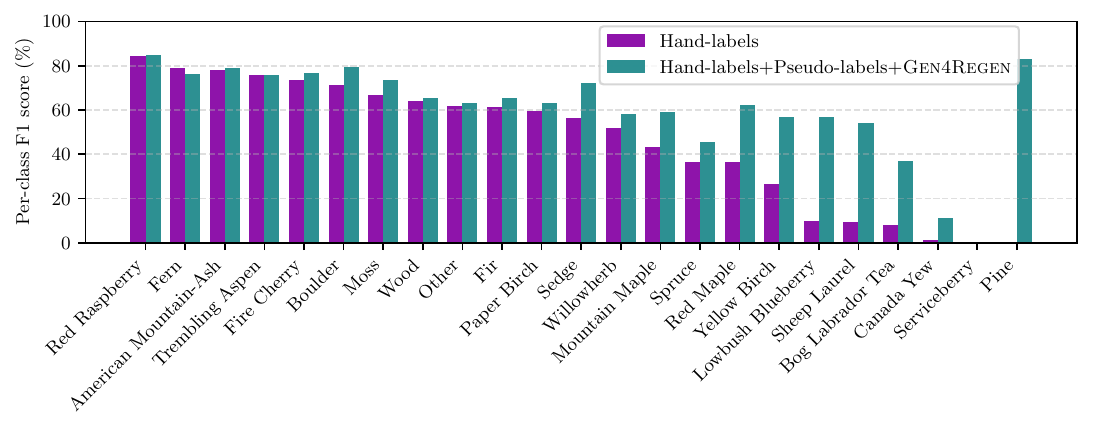}}
    \caption{Comparison of per-class F1 score performance with DINOv3 when training with only hand-labelled data versus our hand-labelled, pseudo-labelled, and \GenAiDataset data.
    The use of additional class labels can greatly improve the F1 score of classes that were previously suffering from class imbalance, while maintaining performance on other classes.}
    \label{fig:per_class_f1}
\end{figure}

Across all metrics, DINOv3 achieves competitive performance; we thus choose this model for further analysis.
The per-class F1 scores, shown in \autoref{fig:per_class_f1}, highlight the main advantage of our proposed methodology: the mitigation of the performance degradation typically associated with long-tail class distributions, as illustrated in \autoref{fig:class_distribution}.
While the baseline DINOv3 model trained exclusively on manual annotations performs adequately on dominant vegetative cover, such as Fern and Red Raspberry, its predictive capacity quickly drops for underrepresented classes on the tail side of the distribution. 
However, integrating the pseudo-labels alongside the synthetic \GenAiDataset data fills a critical data gap for these rare classes.
Notably, classes such as Lowbush Blueberry, Sheep Laurel, and Pine leap from near-zero F1 scores to values exceeding \SI{50}{\%}.
Other classes, such as Bog Labrador Tea, Canada Yew, and Yellow Birch, incur significant gains as well.
Meanwhile, classes such as Fern, Red Raspberry, Fire Cherry, and Trembling Aspen that were already well-segmented generally maintain their performance, indicating that the use of pseudo-labelled and synthetic data is not a trade-off.
Consequently, this multi-source training approach is effective for broad-spectrum species composition mapping in highly imbalanced natural environments.

\begin{figure}[h!]
    \centering{\includegraphics[width=0.92\textwidth]{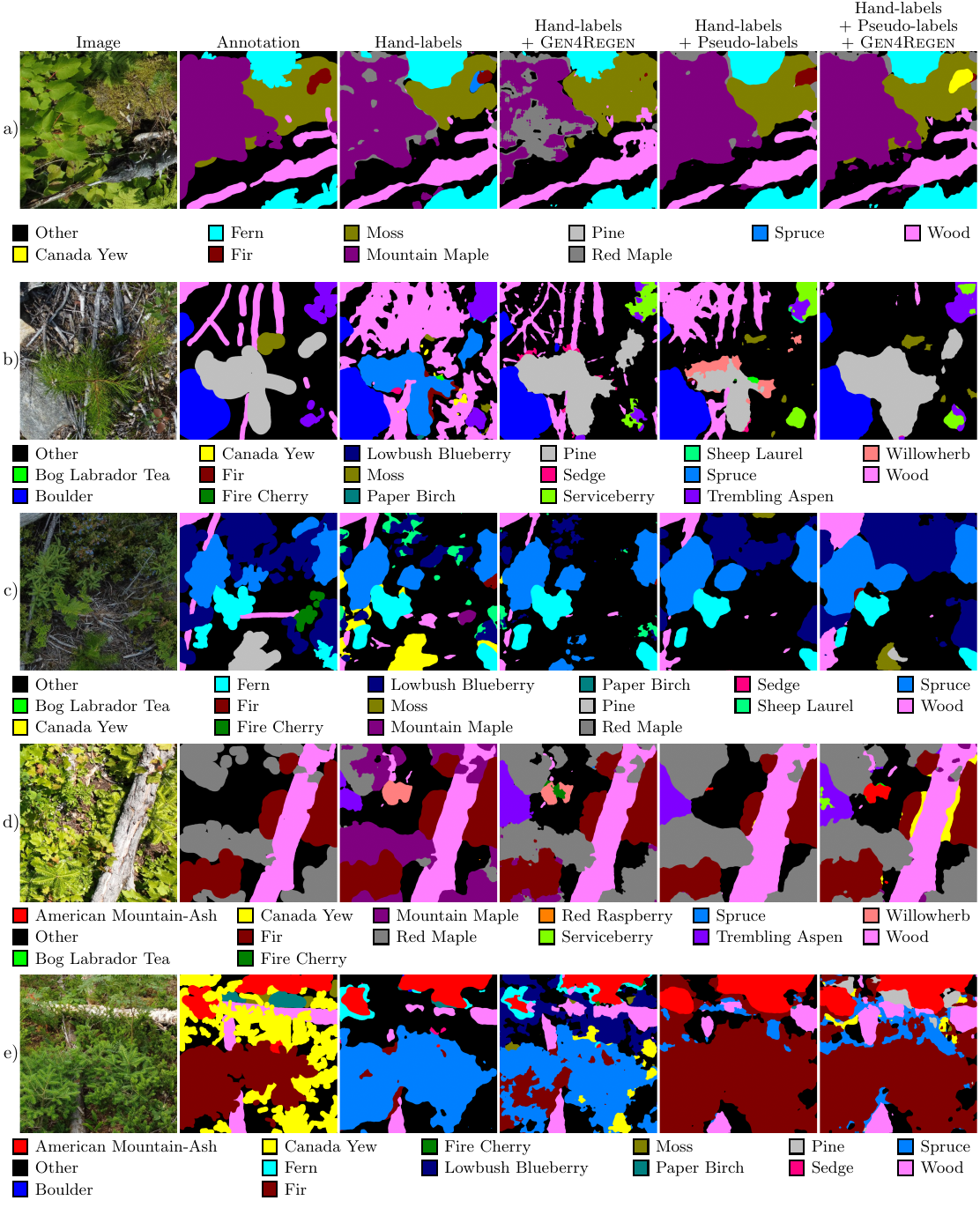}}
    \caption{Selected qualitative examples using the DINOv3 model, for different combinations of training data.}
    \label{fig:qualitative_results}
\end{figure}

\begin{figure}[h!]
    \centering{\includegraphics[width=1.0\textwidth]{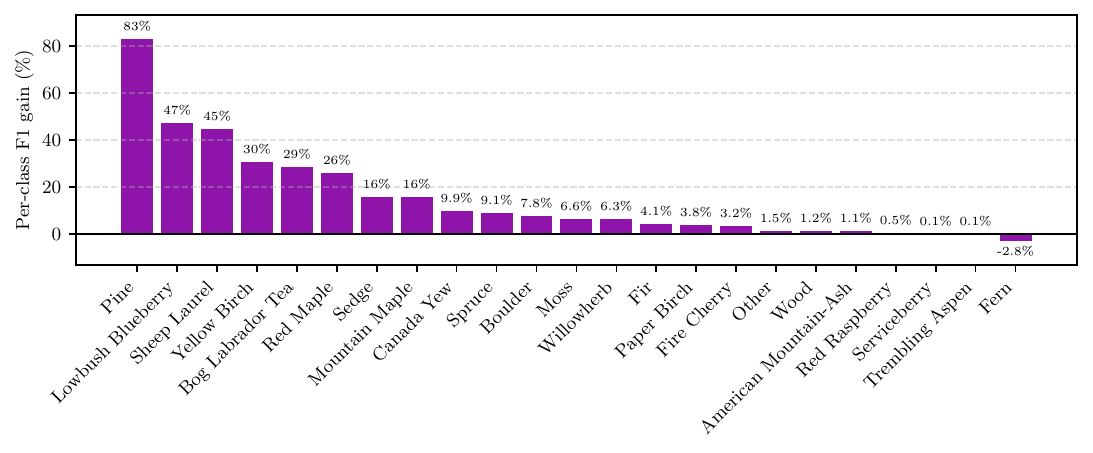}}
    \caption{Comparison of per-class F1 score performance gains with DINOv3 when training with only hand-labelled data versus our hand-labelled, pseudo-labelled, and \GenAiDataset data.
    The use of additional class labels can greatly improve the F1 score of classes that were previously suffering from class imbalance, while maintaining performance on other classes.}
    \label{fig:per_class_gain}
\end{figure}

Qualitative results for the DINOv3 model across different training configurations are presented in \autoref{fig:qualitative_results}.
Notably, models trained with the Labelled + \GenAiDataset configuration produce significantly more intricate segmentation boundaries.
This finer spatial granularity directly reflects the increased mask complexity inherent to \GenAiDataset, as further quantified in \autoref{sec:genai_dataset_analysis}.
A similar, albeit slightly attenuated, level of detail is observable in the Pseudo-Labels + Labelled + \GenAiDataset predictions.
Examining the first row of \autoref{fig:qualitative_results} \textbf{a)}, the addition of pseudo-labels helps the network predict the Mountain Maple class, which was otherwise confused with Red Maple.
The second row \textbf{b)} shows how the integration of AI-generated training data enables the network to correctly delineate Pine, whereas models lacking this synthetic data misclassify it as Spruce.
This visual evidence corroborates \autoref{fig:per_class_gain}, which highlights Pine as the class benefiting most from added data sources.
In the third row \textbf{c)}, the addition of pseudo-labels and \GenAiDataset reduces the confusion between Fir, Spruce, and Canada Yew, three visually similar classes.
In addition, as with the second row, combining both pseudo-labels and \GenAiDataset helps the model predict the Pine class.
In the fourth row \textbf{d)}, pre-training on pseudo-labels significantly reduces confusion of Red Maple with other classes.
Finally, in the fifth row \textbf{e)}, the addition of the pseudo-labels reduces the confusion between Spruce and Fir.
More importantly, only the models trained with AI-generated data successfully detect the presence of Canada Yew.
Since the predicted surface area may be smaller than the coarse ground truth, it inevitably leads to a penalty in pixel-level metrics such as mIoU.
However, from an ecological standpoint, the critical outcome is that the presence of this rare class was successfully detected in the scene.
Importantly, in addition to an overall improvement when adding the pseudo-labels and the \GenAiDataset dataset, \autoref{fig:per_class_gain} shows how the additional data leads to often significant performance gains for all classes except Fern, which sees a very slight decrease of \SI{2.8}{\pp}.

\subsection{Evaluating Performance without Training on Hand-labelled Data}

Further experiments were conducted on DINOv3 to assess performance \textit{without} reliance on any hand-labelled segmentation masks during training.
The results of these experiments are reported in \autoref{tab:results_no_label}.
Remarkably, the DINOv3 model was able to reach an F1 score of \SI{53.25}{\%} when trained exclusively on AI-generated image-mask pairs. 
This empirically demonstrates that a modern image generator possesses the semantic understanding and structural fidelity required to synthesize viable training datasets.
These findings signal a fundamental paradigm shift in semantic segmentation: the traditional bottleneck of manual annotation is being circumvented by scalable, prompt-driven data synthesis.
When trained solely on our pseudo-labels, the same model achieved an F1 score of \SI{48.86}{\%}, while utilizing both data sources yielded \SI{57.27}{\%}.
This compounding performance strongly reaffirms the orthogonality of the synthetic and pseudo-labelled datasets, following the same trend we noted earlier.
It also emphasizes that utilizing vast amounts of real, unlabelled data is still highly desirable for maximizing downstream gains.
For completeness, we have \SI{60.75}{\%} F1 score, \SI{46.42}{\%} mIoU, and \SI{68.20}{\%} pixel accuracy for the DINOv3 model trained on all datasets, including labelled data in \autoref{tab:results}.
These results show that hand-labelled data brings somewhat limited gains, which is the same conclusion as \citet{nasiri_using_2025}.
We conjecture that the pseudo-labels bring useful semantic information but offer very coarse, incorrect masks, as seen in \autoref{sec:genai_dataset_analysis}.
Conversely, the AI-generated images contain less informative visual features, while the corresponding masks are dense and precise.
Consequently, standard quantitative metrics likely exhibit a negative bias against the pure generative AI approach, artificially penalizing the model for achieving higher spatial precision than our coarser ground truth.

\begin{table}[!h]
    \centering
    \caption{Segmentation performance based on the data used without hand-labelled data with DINOv3.
    For comparison, we include our baseline results using all data sources.
    Best results without hand-labels are highlighted in \textbf{bold}.\label{tab:results_no_label}}
    {\begin{tabular}{cccccc} \toprule
    Hand-labels & Pseudo-labels & \GenAiDataset & F1 score & mIoU & Pixel \\
    (\numlabelled images) & (\numunlabelled images) & (\numregen images) & & & Accuracy \\
    \midrule
    -- & \checkmark & -- & 48.86$_{\pm 2.31}$ & 35.20$_{\pm 1.83}$ & 59.12$_{\pm 0.88}$ \\
    -- & -- & \checkmark & 53.25$_{\pm 2.36}$ & 38.50$_{\pm 2.12}$ & 59.74$_{\pm 2.00}$ \\
    -- & \checkmark & \checkmark & \textbf{57.27}$_{\pm 1.25}$ & \textbf{42.33}$_{\pm 1.09}$ & \textbf{63.12}$_{\pm 0.94}$\\
    \midrule
    \checkmark & \checkmark & \checkmark & 60.75$_{\pm 0.69}$ & 46.42$_{\pm 0.78}$ & 68.20$_{\pm 0.51}$\\
    \midrule
    \end{tabular}}{}
\end{table}

\subsection{Testing the Zero-Shot Semantic Segmentation Capabilities of Nano Banana Pro}

Similarly to \citet{gabeur_image_2026}, we evaluated the Nano Banana Pro generative model for zero-shot segmentation on a 60-image subset of our labelled data, carefully chosen to reflect the diversity and complexity of our task.
To this end, we employed several prompting schemes and recovered the prediction masks by post-processing the generated RGB values (\refSection{sec:GenAIgeneration}).
In a strict zero-shot setting, by requesting the image generator to segment all 23 classes without any contextual cues, the model achieved a surprisingly high F1 score of \SI{36.72}{\%}, as shown in \autoref{tab:zero_shot_segmentation}.
This aligns closely with the findings of \citet{gabeur_image_2026}, who demonstrated that a lightly fine-tuned version of Nano Banana Pro achieves state-of-the-art-level performance for zero-shot semantic segmentation on Cityscape \cite{Cordts2016Cityscapes}.
Nevertheless, it falls behind the DINOv3 model trained solely on data generated by the same image generator by \SI{16.53}{\pp} (see \autoref{tab:results_no_label}).
One plausible explanation for this difference lies in the manual curation process applied during the creation of the \GenAiDataset dataset.
In addition, during the generation process, the model is provided with explicit class instructions in a very detailed prompt; however, during zero-shot evaluation, it must discriminate among \numclasses classes, which increases the likelihood of hallucinations.

We subsequently investigated the performance gains achievable by introducing auxiliary cues.
These prompts consisted of restricted class lists derived either from our pseudo-labelling technique (\refSection{sec:pseudolabelgeneration}) or directly from the ground truth, as well as visual cues where the pseudo-label images were provided alongside their respective classes.
For an example of one of our prompts, refer to \autoref{sec:zeroshot_prompt}.
As shown in \autoref{tab:zero_shot_segmentation} (lines 2-4), integrating pseudo-labels as auxiliary inputs serves as a driver for performance improvements, although most of the gains result from reducing the set of classes provided in the prompt.
Despite the image generator's strong generative capabilities, the simple sliding-window pseudo-labelling method proposed by \citet{nasiri_using_2025} independently yields superior segmentation results.
The generative model only outperforms this pseudo-label baseline when explicitly prompted with the exact ground-truth plant composition, an oracle scenario that is unrealistic for operational deployment.
This suggests that while the gap between zero-shot generative models and weakly-transferred supervised approaches is rapidly closing, specialized, domain-adapted methods still maintain a slight but distinct edge.

\begin{table}[!h]
    \centering
    \caption{Performance of Nano Banana Pro in zero-shot semantic segmentation, under different prompt configurations. 
    Evaluation was conducted on a 60-image subset of our labelled dataset, which was carefully chosen to reflect the complexity and diversity of our task. 
    Best results are highlighted in \textbf{bold}.
    For comparison, we include baseline results with the original pseudo-labels on this 60-image subset.
    \label{tab:ai_input}}
    {\begin{tabular}{lcccc} \toprule 
    Input class list in prompt & Pseudo-Labels & F1 score & mIoU & Pixel Acc \\
    \midrule
    Full & -- & 36.72 & 24.76 & 43.67 \\
     & \checkmark & 40.59 & 27.50 & 49.14 \\
    Classes in Pseudo-Labels & -- & 48.27 & 34.03 & 55.58 \\
     & \checkmark & 50.26 & 35.79 & 57.40 \\
    Classes in Annotations & -- & \textbf{51.65} & \textbf{37.81} & \textbf{61.68} \\
    \midrule
    \multicolumn{2}{l}{Baseline with Pseudo-labels} & 50.19 & 36.14 & 57.50 \\
    \midrule
    \label{tab:zero_shot_segmentation}
    \end{tabular}}{}
\end{table}

\subsection{Ablation Study on the Number of \GenAiDataset Images}

\begin{figure}[!h]
    \centering
    \includegraphics[width=0.7\linewidth]{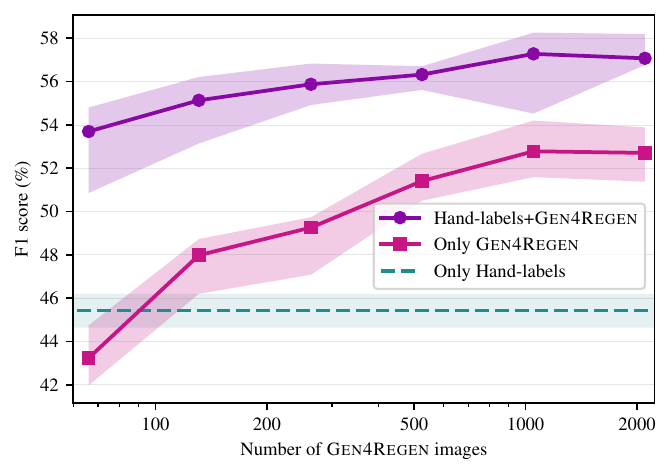}
    \caption{F1 score based on the number of AI-generated images used during training.
    The DINOv3 model was used, and the 40:60 synthetic:real ratio was used for all experiments.
    The \ac{IQR} are based on eight training runs, each with a distinct random seed, that were executed for each of the experiments.
    The x-axis is in log space.
    }
    \label{fig:ablation}
\end{figure}

While our previous experiments utilized the full \GenAiDataset corpus during training, assessing the data efficiency of this generative approach is equally critical.
To quantify the model's sensitivity to synthetic data scaling, we conducted an ablation study by systematically downsampling the number of generated images.
\autoref{fig:ablation} illustrates the resulting F1 scores for the DINOv3 model, replicating the configuration from \autoref{tab:results} (line 11) trained on a mix of manually labelled and synthetic data, across varying synthetic dataset sizes.
The trend reveals a steep initial surge in performance, which subsequently transitions into a near log-linear scaling, yielding an approximate gain of \SI{0.5}{\pp} for each doubling of the synthetic dataset size.
For contextual comparison, \textit{Gen-n-Val} \cite{Huang2025GennValAI} reports an average performance gain of approximately \SI{0.25}{\pp} per dataset doubling for rare class metrics on the mature and exhaustive COCO dataset, using their AI generation approach.
Ultimately, the key takeaway from this ablation is the sample efficiency of the generated data: the integration of even a small number of synthetic images with their annotation masks yields rapid performance gains.
This empirically demonstrates that such an approach is a highly cost-effective strategy for bootstrapping data-scarce domains.
While results obtained using the full dataset are comparable to those with half the data, the observed variance remains non-negligible, leaving open the possibility that additional data could still yield more stable or potentially improved results.

\section{Discussion}

The experimental results presented in this study highlight a novel application of large-scale image generators for complex computer vision tasks in forestry.
While previous techniques for synthetic data generation have lacked photorealism, required specialized models, been limited to simple contexts, or did not address segmentation mask generation, we demonstrate that they can now be used as a dataset generator for model training, generating both images and semantic masks simultaneously.
Specifically, we show that integrating synthetic data into neural network training for our data-scarce and imbalanced context yields a substantial performance increase of \SI{11.77}{\pp} using the DINOv3 model.
However, we note that our best results indicate that there is still room for improvement, as current performance likely does not meet the quality and reliability required for commercial deployment.
We hope that the continuous improvement of image generators will bridge the performance gap needed for real-world applications while reducing reliance on hand-crafted data annotations.
To contextualize our findings, this discussion first traces the evolution of vision-language architectures in vegetation perception and explores the distinct advantages of using generative AI as a source for training data.
We then examine the operational limitations and practical trade-offs of this approach, particularly our reliance on closed-source models and the associated challenges of latency, \ac{API} stability, and scaling costs.

\subsection{The Evolution of Vision Foundation Models in Vegetation Perception}

A recent trend has been to leverage vision foundation models, such as the Segment Anything Model (SAM) \cite{kirillov_segment_2023}, primarily as interchangeable building blocks \cite{duguay2026selvamask} or to refine pseudo-masks in a post-processing stage \cite{Grondin2024}.
In addition, recent work by \citet{ROGGIOLANI2026} explored the use of image generators, adapting models for agricultural plant segmentation, which necessitates tailored refinement to overcome the limitations of off-the-shelf models.
Our evaluation of the Nano Banana Pro model situates this work within the rapidly accelerating adoption of large-scale foundation models for vegetation perception.
By reframing perception as image generation, these models can parse natural language prompts and identify specific morphological traits without requiring explicit prior training on those target classes, allowing new classes to be added dynamically.
Leveraging this generative capability, we introduce a novel workflow that utilizes the Nano Banana Pro model to synthesize a fully paired dataset of synthetic images and pixel-aligned masks specifically optimized for training downstream semantic segmentation networks.
Importantly, our experiments validate this approach, demonstrating that fully generated datasets can significantly improve semantic segmentation performance and, in some cases, outperform smaller manually annotated datasets.

\subsection{Advantages of Generative AI for Semantic Segmentation Training Data Generation}
While commercial generative vision models enable impressive zero-shot semantic segmentation, their true value in our context lies in their flexibility and capacity for data synthesis, without requiring additional training.
Utilizing generative AI as a source for training data provides several distinct advantages:
\begin{itemize}[topsep=0pt, labelsep=1em, leftmargin=*]
\item[] \textit{Accelerated Project Lifecycles}: For forestry drone image analysis, practitioners no longer need to wait for specific data-gathering seasons to build, test, and validate their training pipelines.
Large quantities of finely labelled data can be generated immediately.
Our results even show that models trained on a large enough corpus of AI-generated data can outperform those trained on smaller manually labelled datasets.
\item[] \textit{Dataset Balancing and Diversity}: Creating structural and visual diversity, such as raindrops, varied lighting, and seasonal changes such as blooming, is far easier to prompt synthetically than to capture in the wild.
This targeted generation actively rebalances the dataset, resulting in a higher overall model performance by ensuring better representation of rare classes.
Furthermore, natural environments are inherently unbalanced, making dataset balancing through \textit{in situ} data acquisition particularly challenging.
\item[] \textit{Orthogonal Generalization}: As demonstrated in our results, Generative AI acts orthogonally to pseudo-labelling.
Because the synthetic domain differs significantly from real images, injecting this variety forces the network to learn stronger generalized features, better preparing the system to handle images from new geographical localities during operational inference.
\end{itemize}

\subsection{Operational Limitations and Practical Trade-offs}

Despite these theoretical and developmental advantages, utilizing proprietary generative models as direct substitutes for trained specialist networks for generating training data or zero-shot inference introduces several operational constraints:
\begin{itemize}[topsep=0pt, labelsep=1em, leftmargin=*]
\item[] \textit{Computational Latency and Edge Deployment}: In our pipeline, generating a segmentation mask required approximately 10 to 30 seconds per image, compounded by network connection variability.
Even when mitigated through batch-requesting, this is orders of magnitude slower than the roughly 200 milliseconds required by a locally trained specialist network.
Furthermore, the closed-source nature and prohibitive computational requirements of these large-scale models make deployment on edge devices, such as \ac{UAV} onboard computers, currently unfeasible.
\item[] \textit{API Dependence and Reproducibility}: 
Relying on remote, closed-source commercial \acp{API} introduces long-term uncertainty. 
Model providers may alter internal weights (leading to variations even when re-downloading the "same" image), deploy distilled versions to reduce server costs, or deprecate models entirely, as exemplified by OpenAI's Sora retirement \cite{openai_sora_2024}.
Consequently, previously successful prompt formulations may unexpectedly lose their effectiveness.
In our experience, the \texttt{gemini-3-pro-image-preview}, available from November 20, 2025, to July 17, 2026, which we used for the \GenAiDataset dataset creation, had a generation success of around \SI{85}{\%}.
In striking contrast, the more recent \texttt{gemini-3-pro-image}, released on May 28, 2026, has a generation success of under \SI{25}{\%}.
Nonetheless, image generators are becoming more powerful, and newer image generators are likely to be more reliable.
\item[] \textit{Reliability and Hallucinations}: The probabilistic nature of generative architectures can lead to output hallucinations and a highly bimodal performance curve (either working perfectly or failing entirely).
In our evaluation, 10 to \SI{15}{\%} of the zero-shot segmentation predictions contained errors that required outright discarding to maintain downstream system integrity.
While test-time augmentation (e.g., querying the model with slightly rotated inputs) might mitigate stochastic errors, the generation remains highly sensitive to prompt formulation.
We showcase a few failure cases of conjoint image and mask generation in \autoref{sec:generation_failure_cases}.
\item[] \textit{Prohibitive Scaling Costs}: Contrary to \citet{ROGGIOLANI2026}, we did not investigate the potential of using the pseudo-masks obtained from the image generator for training purposes.
While compelling, using a commercial image generator over \num{550000} unlabelled \SI{1}{MP} images would result in a prohibitive cost: at the current pricing model, the estimated expense would be over \num{55000} USD.
Moreover, sliding-window pseudo-labelling techniques \cite{nasiri_using_2025,soltani_simple_2024} already provide excellent results, outperforming zero-shot segmentation, as shown in \autoref{tab:zero_shot_segmentation}.
\item[] \textit{Phenotypic Constraints}: Although generative AI offers a powerful offline mechanism for synthesizing training data, its application is currently constrained by the model's internal phenotypic priors, which may restrict the actual structural and visual diversity of the plants generated.
As shown in \autoref{fig:ablation}, generating an immense corpus of AI-generated data might not necessarily enable stronger generalization when training on this dataset.
\item[] \textit{Legal Restrictions and Terms of Service}: Practitioners must navigate  commercial terms of service.
Extracting training masks from proprietary model outputs might conflict with rules that explicitly prohibit model distillation or the training of downstream competitive networks.
\end{itemize}

Consequently, while these image generators are excellent tools for dataset augmentation, operational forestry monitoring still strictly requires locally trained supervised networks to ensure low latency, the ability to deploy on edge devices, and reliable output quality.
However, the continuous and rapid improvement in image generator capabilities dictates that these large-scale models be factored into the strategic roadmaps for future ecological monitoring systems relying on semantic segmentation.
Ultimately, this study serves to highlight the transformative potential of generative data synthesis, encouraging its immediate integration into the development lifecycles of specialized ecological \ac{AI}.

\section{Conclusion}

In this study, we explored the role that generative \ac{AI} can play in addressing the critical bottleneck of data scarcity and severe class imbalance in high-resolution semantic segmentation, using forest regeneration monitoring as a test case.
To this effect, we introduced a comprehensive methodology leveraging an expanded dataset of over \num{25000} UAV images (\ourdataset) and a novel synthetic dataset (\GenAiDataset) comprising \num{2101} photorealistic image-mask pairs generated via the Nano Banana Pro model, part of Google's Gemini ecosystem.
Our evaluations using Mask2Former, DINOv2, and DINOv3 architectures demonstrate that generative AI not only helps alleviate the need for expert manual annotation but is also highly orthogonal and complementary to other data sources.
Integrating these synthetic and pseudo-labelled datasets yields substantial performance improvements, exceeding a \SI{15}{\pp} gain in F1 score, with most of the gains for underrepresented, rare classes.
Consequently, the ability to generate training data via prompts offers an affordable way to rectify class imbalances and domain gaps.
We also demonstrate that Nano Banana Pro, although not explicitly trained for plant identification, exhibits surprising zero-shot semantic segmentation capabilities in this complex natural environment, with an F1 score of \SI{36.72}{\%} over 23 classes, while a network trained solely on AI-generated images reaches an F1 score of \SI{53.25}{\%}.

Our findings demonstrate that by jointly generating high-fidelity, photorealistic images and their corresponding semantic masks, image generators are now a viable source of training data.
This approach drastically reduces the manual annotation burden, in particular for complex morphological structures; while humans can provide coarser, high-level class verification, the generative model natively captures the complex spatial boundaries of plant contours.
Moreover, practitioners can now rapidly prototype perception pipelines without needing to collect a single real-world image, decoupling system development from seasonal constraints.
Thus, a forestry project can be fully initialized during the winter; once drone data acquisition begins in the spring or summer, initial classification models are already operational to guide further collection and labelling efforts.
We believe that large-scale generative models will soon be capable of acting as agile training data factories, enabling the rapid and cost-effective development of perception systems specialized for niche applications.

For future work, we will expand along several promising avenues.
First, building upon the findings of \citet{ROGGIOLANI2026}, we aim to investigate whether zero-shot segmentation masks can serve as complementary signals to our pseudo-labels during the pre-training of vision backbones.
Second, because Nano Banana Pro supports multimodal inputs, we plan to incorporate reference images from iNaturalist directly into the generative prompting process.
Leveraging the vast geographical diversity of iNaturalist observations could significantly expand the phenotypical variety of the synthesized data, further enhancing the model's capacity to generalize across distinct geographic regions.
Furthermore, we will investigate the application of generative AI for semantic-driven data augmentation, such as altering the illumination of real \ac{UAV} images while strictly preserving their hand-labelled ground-truth masks.
Finally, given that our \GenAiDataset currently contains only \numregen images, we aim to expand this corpus size to identify the performance ceiling of synthetic training data.
In tandem, we will explore domain adaptation frameworks, such as Domain Adversarial Neural Networks \cite{DANN2016}, to actively mitigate the domain gap between synthetic and real-world images.

\section{Acknowledgements}

We sincerely thank the Ministry of Natural Resources and Forests of Quebec for their support throughout the project. 
We would also like to thank Justine Therrien and Julien St-Louis, whose contributions were essential to the annotation process, as well as Benjamin Jeanrie-Chouinard for his significant contribution to data collection.
We also thank the FORAC Research Consortium for their guidance.

\section{Funding}

This research was supported by the Ministry of Natural Resources and Forests [3329-2022-2204-1]; the Natural Sciences and Engineering Research Council of Canada Discovery Grant [RGPIN-2022-04741]; the Canada Foundation for Innovation Fund [39709, PI: E. Thiffault and F. Anctil]; and the Canada Graduate Scholarship-Master's Program.

\bibliographystyle{foresj}
\bibliography{references}

\clearpage
\newpage
\appendix
\section{Confusion Matrices}

Generating images can help mitigate model confusion, even between plant classes which aren't directly addressed.
For example, we noted that even if we added only a few images with Yellow Birch and Paper Birch, these classes were significantly less confused, hinting at the model's better understanding of general phenotypes.
In addition, Red Maple and Mountain Maple are also less confused with the addition of pseudo-labels and the \GenAiDataset dataset.
Importantly, the model performs much better on rare classes that were rarely predicted when trained only on labelled data, as shown by the four classes (Bog Labrador Tea, Lowbush Blueberry, Pine, and Sheep Laurel), whose correct prediction rates increased from under \SI{10}{\%} to over \SI{38}{\%}.
This shows how the additional data can greatly improve model performance on classes in the long-tail distribution.

\begin{figure}[!h]
    \centering
    \includegraphics[width=0.9\textwidth]{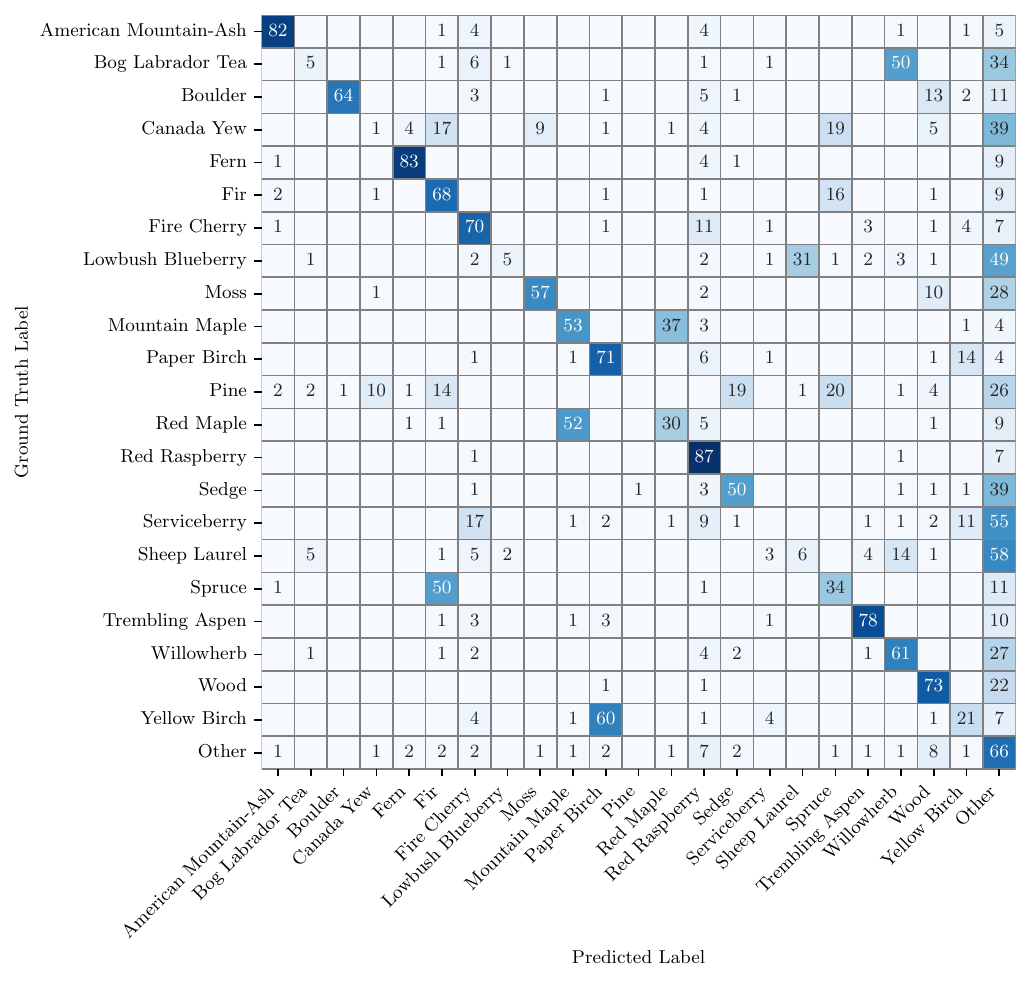}
    \caption{Confusion matrix for DINOv3 trained with the hand-labelled data.
    }
    \label{fig:dino_label_cm}
\end{figure}

\begin{figure}[!h]
    \centering
    \includegraphics[width=0.9\textwidth]{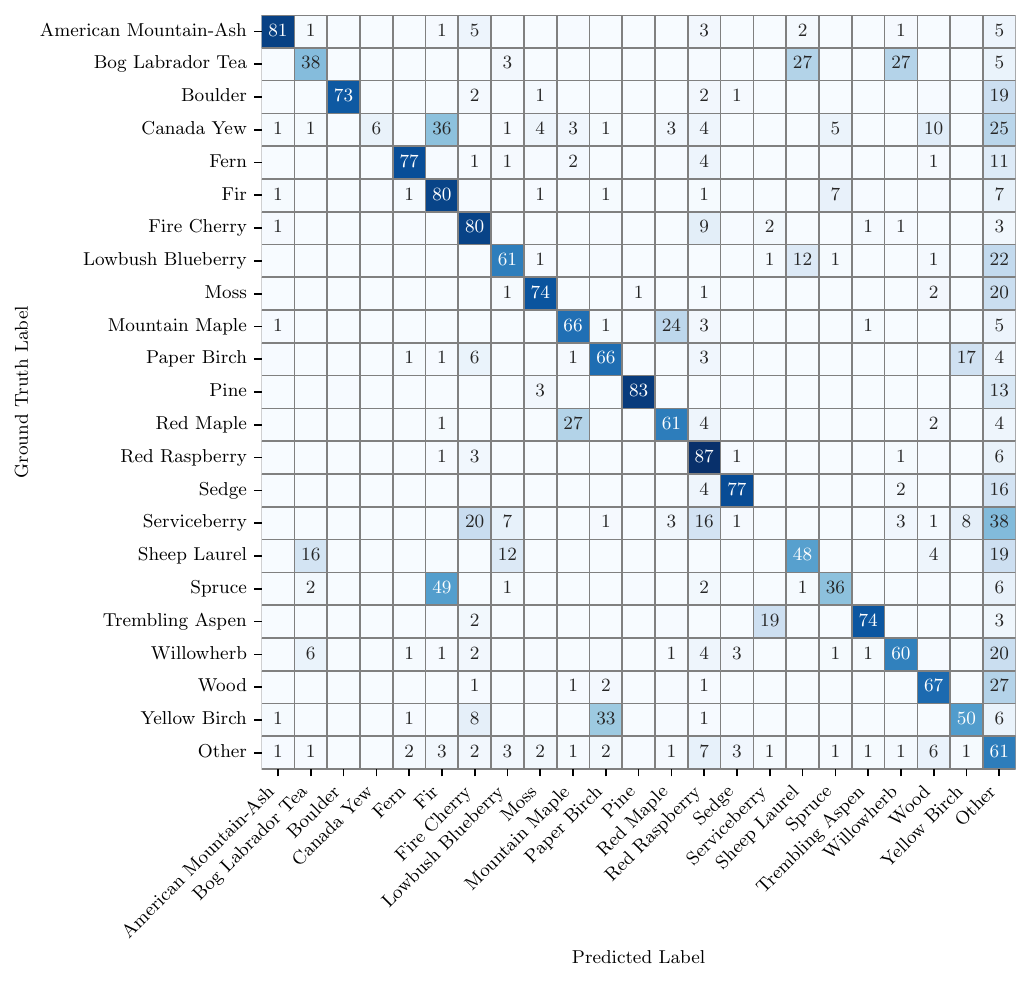}
    \caption{Confusion matrix for DINOv3 pre-trained on the pseudo-labels and fine-tuned with the hand-labelled data and \GenAiDataset dataset.
    }
    \label{fig:dino_pl_label_genai_cm}
\end{figure}

\clearpage
\newpage
\section{Per-Class F1 Score}

\begin{figure}[!h]
    \centering
    \includegraphics[width=1.0\textwidth]{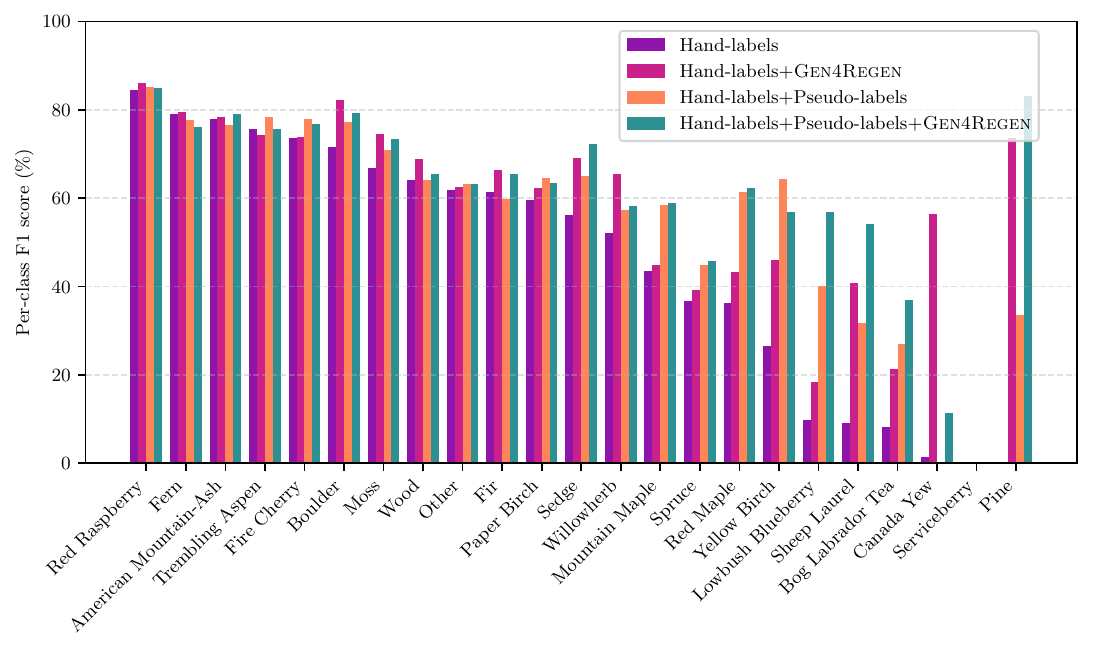}
    \caption{Per-class F1 score for all DINOv3 experiments from \autoref{tab:results}.
    }
    \label{fig:per_class_gain_full}
\end{figure}

\autoref{fig:per_class_f1} shows the per-class F1 score for the DINOv3 model when trained only on hand-labelled data versus when trained on labelled, pseudo-label and \GenAiDataset data.
\autoref{fig:per_class_gain_full} presents the same results, but for all DINOv3 experiments presented in \autoref{tab:results}, allowing finer analysis of the individual contributions of the pseudo-label pre-training and the addition of the \GenAiDataset dataset.

\clearpage
\newpage

\section{\GenAiDataset Dataset Analysis}
\label{sec:genai_dataset_analysis}

Although deep learning approaches can benefit from synthetic data, they must be employed carefully, as there can be an important domain gap between data sources.
To properly analyze this gap, we compare our real and synthetic images with two embedding-based metrics.
The first is \ac{FID} \cite{heusel_2017_fid}, a statistical metric which compares image sets based on mean and covariance matrices.
These matrices are calculated on image features extracted with a neural network.
The second metric is the more recent \ac{CMMD} \cite{jayasumana_2024_cmmd}, similar in concept to \ac{FID} but better aligned with the human perception of image quality, as well as more reliable on smaller datasets.
As a baseline, we compute these metrics between our hand-labelled and unlabelled images, with a lower value indicating higher similarity. 
We computed the metric between the hand-labelled images and the synthetic images, as well as between the hand-labelled images and the iNaturalist subset.
The results are shown in \autoref{tab:distance_metrics} and are indicative of a considerable gap between our real and synthetic images, but still show a closer similarity between those two datasets than between the hand-labelled and iNaturalist subsets.
The relative distance between the datasets is also presented, as the values are elevated due to the limited size of the reference set, especially for \ac{FID}, which is less sample-efficient compared to \ac{CMMD}. 
Importantly, there is a fine line which must be maintained between domain overlap and increased diversity.
Introducing data that strays too far from our real data could negatively impact model training \cite{xu_2023_imperfect}.
This challenge also concerns the synthetic masks, which are significantly more precise than our hand-labelled masks.
This misalignment may introduce discrepancies during training, which could in turn confuse the model.

\begin{table}[!h]
    \centering
    \caption{Perceptual distance metrics between our hand-labelled data and other sets. 
    The relative difference to the hand-labelled and unlabelled data distance metric is also presented. 
    The lower the number $(\downarrow)$, the closer both datasets are, and hence, the trained models are more likely to adapt well to those datasets.}
    \begin{tabular}{l c c c}
        \hline
        & Unlabelled & \GenAiDataset & iNaturalist \\
        \hline
        $(\downarrow)$ FID & \num[round-mode=places, round-precision=2]{74.4808} & \num[round-mode=places, round-precision=2]{168.4454} & \num[round-mode=places, round-precision=2]{186.4194} \\
        $(\downarrow)$ FID (relative to Unlabelled) & - & +\SI{126}{\pp} & +\SI{150}{\pp} \\
        $(\downarrow)$ CMMD & \num[round-mode=places, round-precision=2]{0.708} & \num[round-mode=places, round-precision=2]{1.294} & \num[round-mode=places, round-precision=2]{2.003} \\
        $(\downarrow)$ CMMD (relative to Unlabelled) & - & +\SI{83}{\pp} & +\SI{183}{\pp} \\
        \hline
    \end{tabular}
    \label{tab:distance_metrics}
\end{table}

Thus, in addition to comparing the image features in the \GenAiDataset dataset to real ones, it is also important to evaluate the quality of the generated masks.
Following \citet{Xie2025mass13k}, we use the \ac{mIPQ} metric \cite{osserman_isoperimetric_1978} to measure the overall complexity of masks in our different datasets.
The \ac{mIPQ} is defined as : 
$$mIPQ = {1 \over 4 \pi n} \sum_{i=0}^{n} {L_i^2 \over A_i}, $$
where $L_i$ and $A_i$ denote the mask perimeter and the region area for the $i^{th}$ class, and $n$ denotes the total number of classes in this image.
Importantly, the higher the \ac{mIPQ}, the more complex the mask is.
The results for the \ac{mIPQ} are presented in \autoref{tab:mask_complexity}.
It shows how coarse the pseudo-annotations created by the sliding window approach are and how detailed the masks in the \GenAiDataset are.
The manual annotations are between the two datasets, but still show a significant degree of coarseness.
These values are in line with \citet{Xie2025mass13k}, with their semantic segmentation dataset designed for precise spatial masks exhibiting an mIPQ value of $383 \pm 818$.
In essence, these metrics quantify what each data source brings to model training.
While pseudo-labels offer broad semantic coverage and manual labels provide verified ground-truth reliability, both methods inherently yield coarse segmentation boundaries.
Specifically, the pseudo-labelling algorithm described in \refSection{sec:pseudolabelgeneration} introduces spatial blurring due to its $224 \times 224$ sliding window classification mechanism, whereas manual annotation is fundamentally constrained by the prohibitive labour costs associated with pixel-perfect delineation.
The \GenAiDataset overcomes these limitations by injecting the high-fidelity spatial complexity required to train models on the intricate, interwoven geometries of real vegetative cover.

\begin{table}[!h]
\centering
\caption{Comparison of mask complexity for different datasets.
The mIPQ values are reported as the average over the number of masks used for evaluation, $\pm$ standard deviation.
The higher the mIPQ value, the more complex the masks.}
\label{tab:mask_complexity}
\begin{tabular}{lcc}
\toprule
Dataset & Number of masks & mIPQ \\ \midrule
Hand-Labels & \numlabelled & 9.6$_{\pm 6.1}$ \\
Pseudo-labels & \numunlabelled & 3.4$_{\pm 1.5}$ \\
\GenAiDataset & \numregen & 109.8$_{\pm 90.4}$ \\ \bottomrule
\end{tabular}
\end{table}

\newpage

\section{Prompt Generation Technique}
\label{sec:PromptGenerationTechnique}

We use the following base prompt as a template to generate the annotated images. The elements indicated by \texttt{\textless XXX\textgreater} are replaced by different predefined values to control several aspects of the generated scenes.

\begin{figure}[H]
    \centering
    \includegraphics[width=0.95\textwidth]{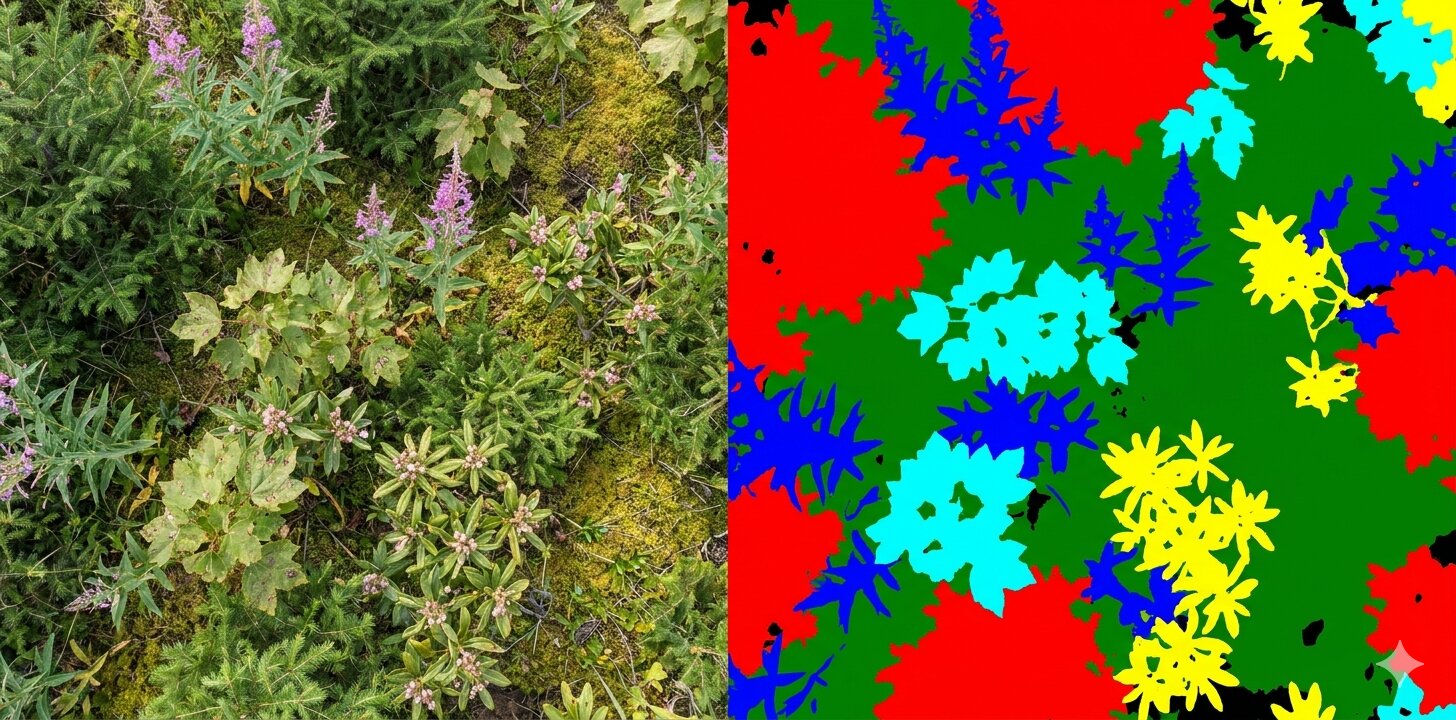}
    \caption{Image generated with Nano Banana Pro using the following variable values:}
    \begin{Verbatim}[breaklines=true, breakanywhere=true,
      breaksymbolleft={},
      breaksymbolright={}]
<DISTURBANCE_EVENTS>                Harvested
<DISTURBANCE_YEARS>                 3 years ago
<PLANT_CLASSES>                     Distributed patches of Abies, small patches of Acer rubrum, 
                                    small patches of Epilobium, and several Kalmia angustifolia
<GLOBAL_DENSITY_DESCRIPTION>        Very dense overlapping vegetation with almost no visible ground
<OTHER_CLASSES>                     Moss covers parts of the ground with green to yellow-green tones.
<GROUND_TEXTURES>                   Uneven terrain with micro-relief variations
<PLANT_STRESS>                      Slightly wilted leaves
<LEAF_VARIATION>                    Some noticeably larger leaves among smaller ones
<FLOWER_DETAILS>                    A few small flowers are visible among the vegetation.
<SEASON_DESCRIPTION>                Mid-summer
<DRYNESS_DESCRIPTION>               Very dry
<RECENT_RAIN_TIMING>                No recent rain
<LIGHT_DESCRIPTION>                 Very sunny morning light.
<PLANT_CLASSES_MASK_COLORS>         Abies : red, Acer rubrum : cyan, Epilobium : blue, 
                                    Kalmia angustifolia : yellow 
<OTHER_CLASSES_MASK_COLORS>         Moss : green
    \end{Verbatim}
    \label{fig:generation_example_summer}
\end{figure}

\begin{figure}[H]
    \centering
    \includegraphics[width=0.95\textwidth]{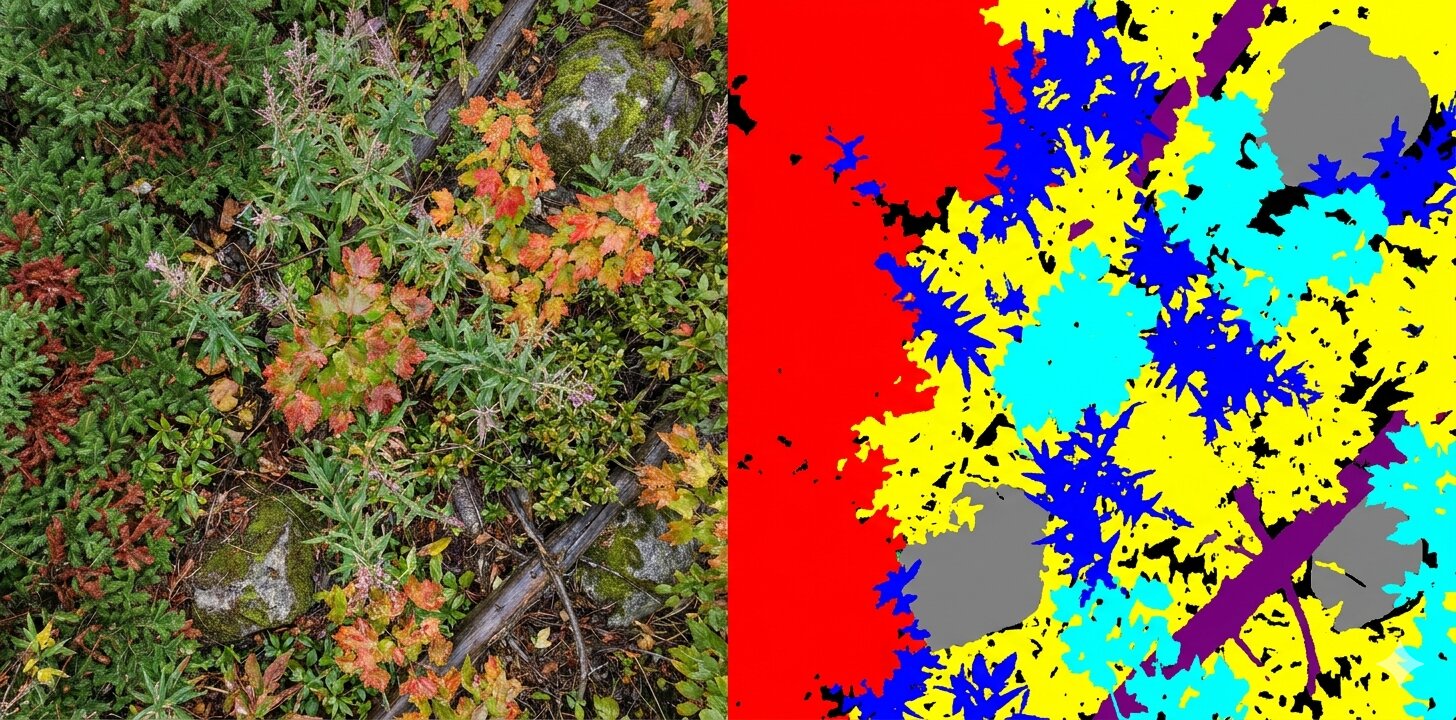}
    \caption{Image generated with Nano Banana Pro using the following variable values:}
    \begin{Verbatim}[breaklines=true, breakanywhere=true,
      breaksymbolleft={},
      breaksymbolright={}]
<DISTURBANCE_EVENTS>                Windthrow
<DISTURBANCE_YEARS>                 6 years ago
<PLANT_CLASSES>                     Noticeable patches of Abies, several Acer rubrum,
                                    some Epilobium, and mixed clusters of Kalmia angustifolia
<GLOBAL_DENSITY_DESCRIPTION>        Very dense overlapping vegetation with almost no visible ground
<OTHER_CLASSES>                     Boulders and dead wood.
<GROUND_TEXTURES>                   Small depressions filled with organic debris
<PLANT_STRESS>                      Mixed healthy and stressed plants
<LEAF_VARIATION>                    Overlapping leaves of different sizes
<FLOWER_DETAILS>                    None
<SEASON_DESCRIPTION>                Mid-fall
<DRYNESS_DESCRIPTION>               Very wet
<RECENT_RAIN_TIMING>                Rain earlier in the day
<LIGHT_DESCRIPTION>                 Overcast
<PLANT_CLASSES_MASK_COLORS>         Abies : red, Acer rubrum : cyan, Epilobium : blue, 
                                    Kalmia angustifolia : yellow
<OTHER_CLASSES_MASK_COLORS>         Wood : purple, Boulders : grey                       
    \end{Verbatim}
    \label{fig:generation_example_fall}
\end{figure}

\begin{figure}[H]
    \centering
    \includegraphics[width=0.95\textwidth]{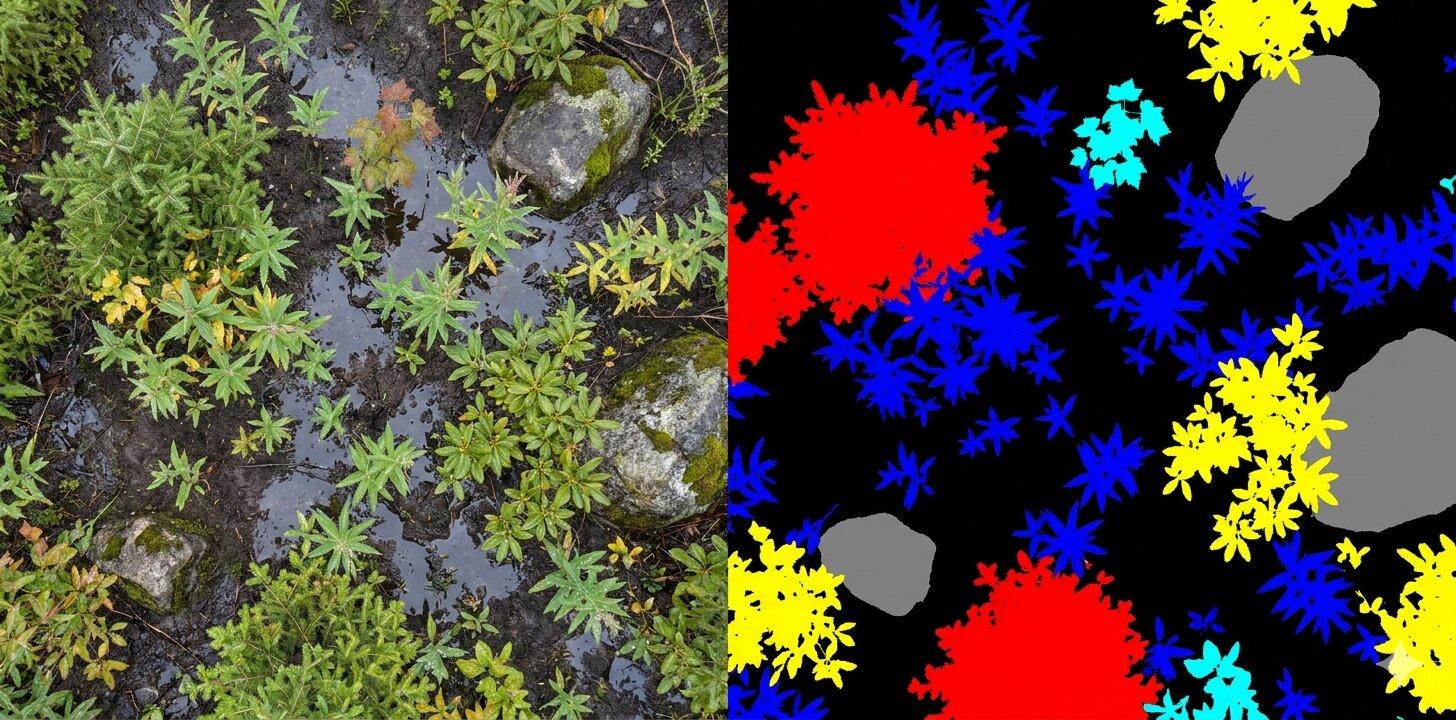}
    \caption{Image generated with Nano Banana Pro using the following variable values:}
    \begin{Verbatim}[breaklines=true, breakanywhere=true,
      breaksymbolleft={},
      breaksymbolright={}]
<DISTURBANCE_EVENTS>                Wildfire
<DISTURBANCE_YEARS>                 2 years ago
<PLANT_CLASSES>                     Small patches of Abies, a few Acer rubrum,
                                    a large number of Epilobium, and several Kalmia angustifolia
<GLOBAL_DENSITY_DESCRIPTION>        Moderately dense vegetation with some visible gaps.
<OTHER_CLASSES>                     Boulders are visible.
<GROUND_TEXTURES>                   Irregular wet areas with darker soil.
<PLANT_STRESS>                      Some yellowing foliage
<LEAF_VARIATION>                    Dense clusters of small leaves.
<FLOWER_DETAILS>                    None
<SEASON_DESCRIPTION>                Early spring
<DRYNESS_DESCRIPTION>               Very wet
<RECENT_RAIN_TIMING>                Rain has just stopped
<LIGHT_DESCRIPTION>                 Partially overcast
<PLANT_CLASSES_MASK_COLORS>         Abies : red, Acer rubrum : cyan, Epilobium : blue, 
                                    Kalmia angustifolia : yellow 
<OTHER_CLASSES_MASK_COLORS>         Boulders : grey     
    \end{Verbatim}
    \label{fig:generation_example_spring}
\end{figure}

In particular, we vary the plant classes as well as the other classes (e.g., moss, boulder, or wood). We also vary the environmental conditions, such as the season, illumination, dryness level, and the time elapsed since the last rainfall. In addition, the appearance of the vegetation is controlled through variables such as plant stress, leaf details, the presence or absence of flowers, and ground textures. Finally, we vary the disturbance event responsible for the regeneration context, as well as the number of years since it occurred. This allows us to generate a wide variety of prompts and, consequently, a diverse set of images. Examples of generated images and their corresponding variable values are shown in Figures~\ref{fig:generation_example_summer}, \ref{fig:generation_example_fall}, and \ref{fig:generation_example_spring}.

\begin{Verbatim}[breaklines=true, breakanywhere=true,
      breaksymbolleft={},
      breaksymbolright={}]
### SCENE DESCRIPTION ###
Split-screen image of 4K resolution, 2:1 aspect ratio.
Left side is a photorealistic high-resolution nadir drone photography flying at a very low altitude of 2 meters of a forest regeneration zone taken with a camera with a focal length of 50 mm. The view is strictly top-down, not oblique. The ground cover is typical of a Quebec forest that was <DISTURBANCE_EVENTS>, <DISTURBANCE_YEARS>.
It contains <PLANT_CLASSES>. <GLOBAL_DENSITY_DESCRIPTION>. 
Fine details such as leaf shapes, veins and small branches are clearly visible. The drone flies low enough to resolve individual leaves with high spatial precision.
<OTHER_CLASSES>. <GROUND_TEXTURES>. <PLANT_STRESS>. <LEAF_VARIATION>. <FLOWER_DETAILS>
<SEASON_DESCRIPTION>, <DRYNESS_DESCRIPTION>, <RECENT_RAIN_TIMING>. <LIGHT_DESCRIPTION>

### VISUAL CONSTRAINTS ###
Photorealistic rendering. Natural colors. No artificial patterns. No repetition artifacts. High-frequency details consistent with real vegetation. No stylization.

### SEGMENTATION MASK ###
The right side shows the exact pixel-aligned semantic segmentation mask of the left image. Each pixel belongs to exactly one class. All objects visible in the left image must appear in the mask with no omissions. The mask uses flat solid colours only, with no shading, gradients, transparency or texture. Boundaries between classes are sharp and perfectly defined. This mask is to be used to train a deep neural network for semantic segmentation with <PLANT_CLASSES_MASK_COLORS>, <OTHER_CLASSES_MASK_COLORS>.
    \end{Verbatim}

\newpage

\newpage
\subsection{Motivation for the choice of Nano Banana Pro}

\begin{figure}[htbp]
    \centering
    \begin{subfigure}[t]{0.48\textwidth}
        \centering
        \includegraphics[width=0.95\textwidth]{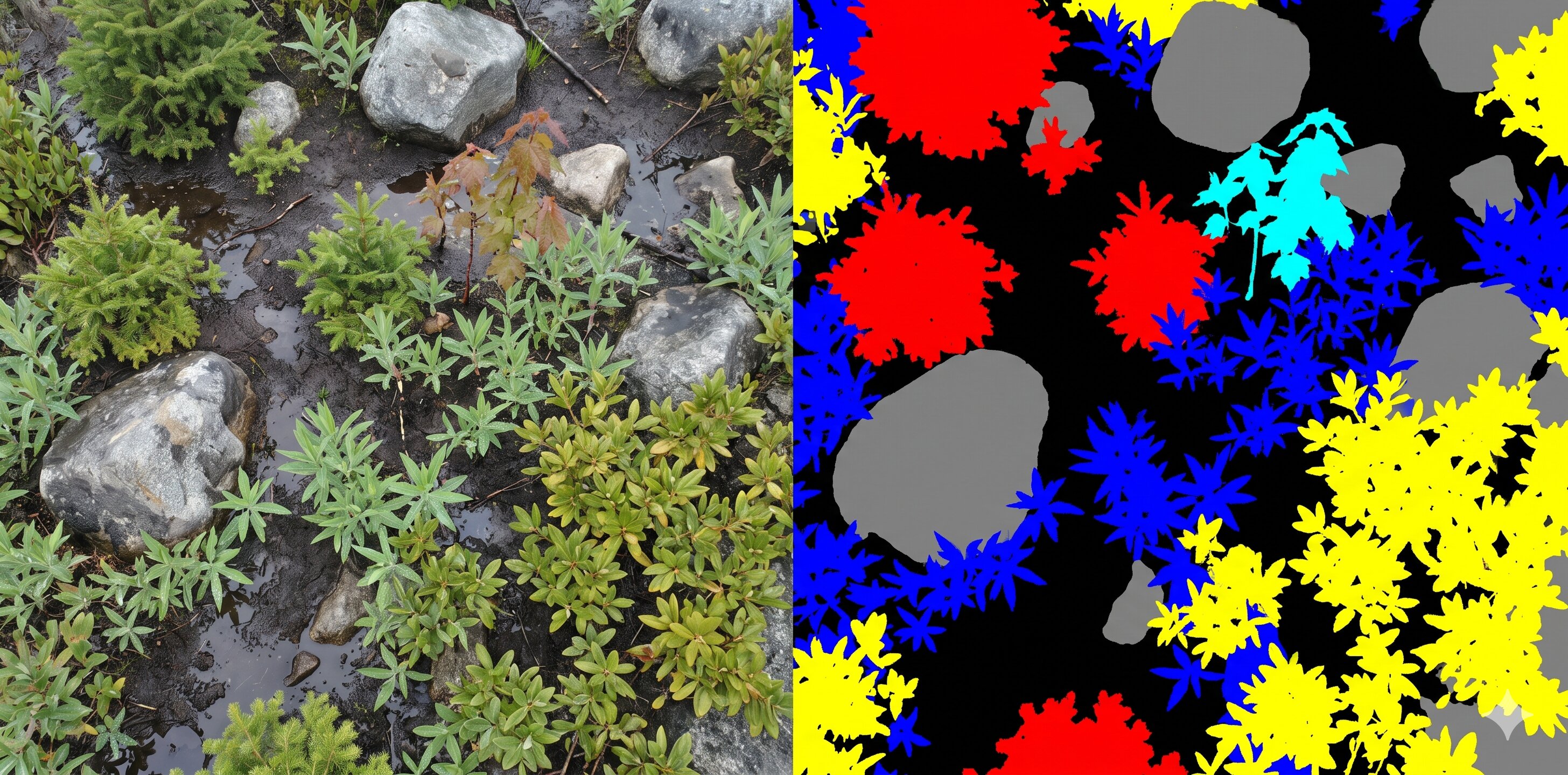}
        \caption{Nano Banana Pro: Generated with the same prompt found in \autoref{fig:generation_example_spring}, but leading to different results.}
        \label{fig:sub1}
    \end{subfigure}
    \hfill
    \begin{subfigure}[t]{0.48\textwidth}
        \centering
        \includegraphics[width=0.95\textwidth]{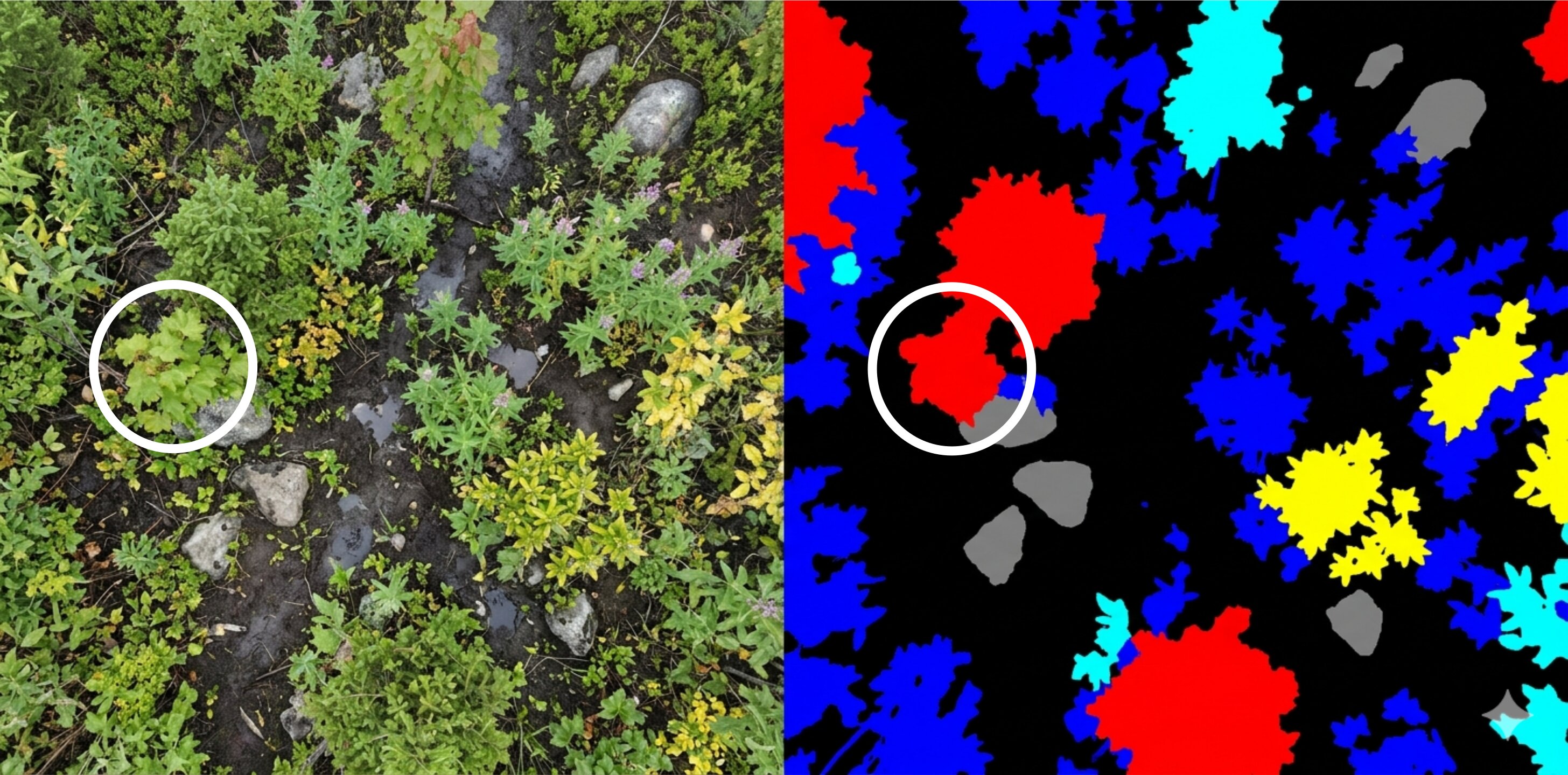}
        \caption{Nano Banana 2: Coarser segmentation masks and clear annotation errors, such as the circled region, which is incorrectly labelled as Fir.}
        \label{fig:sub2}
    \end{subfigure}
    \begin{subfigure}[t]{0.48\textwidth}
        \centering
        \includegraphics[width=0.95\textwidth]{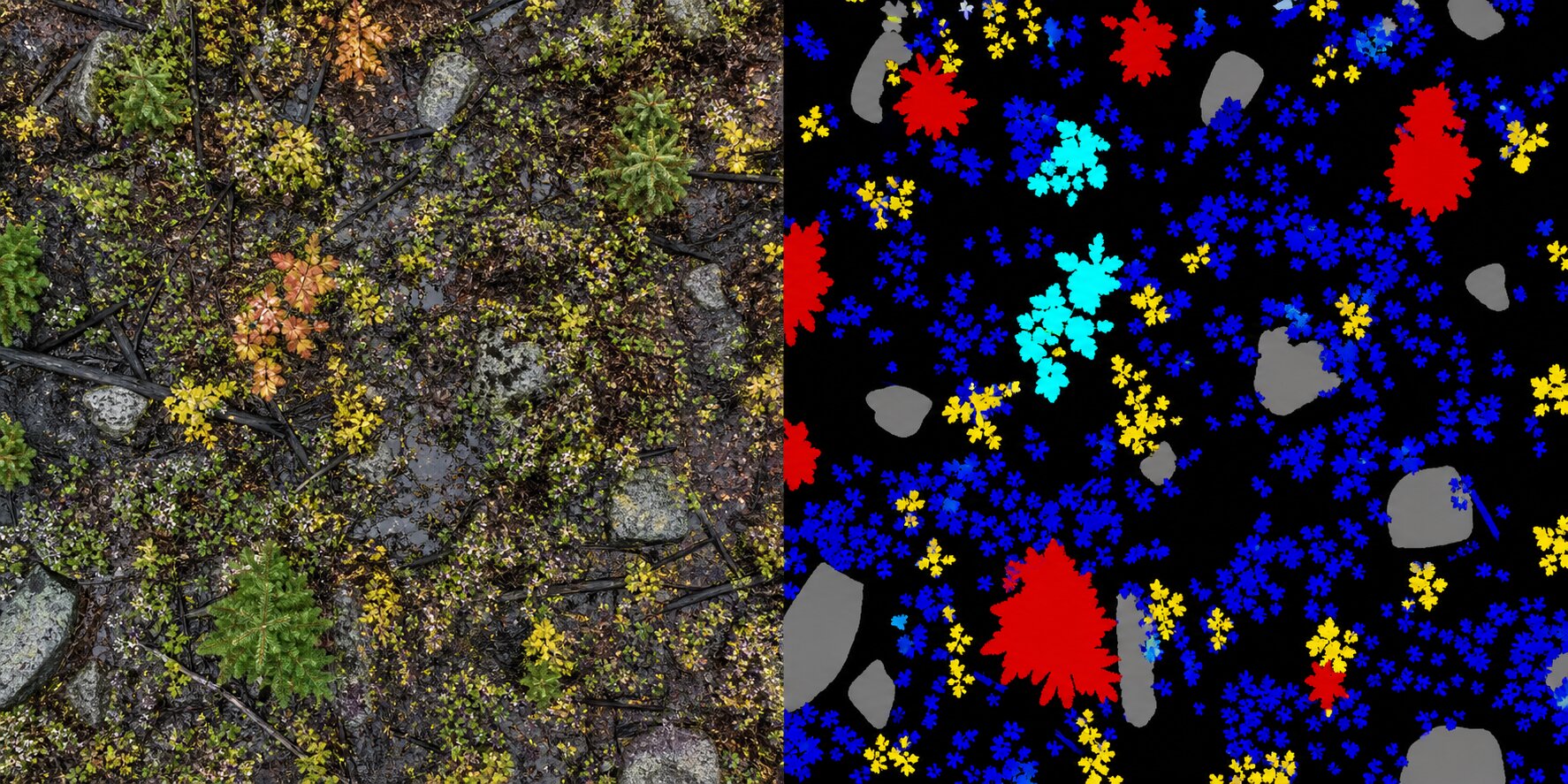}
        \caption{GPT Image 2: Generated images tend to share a similar appearance, with many small plants scattered across the forest floor.}
        \label{fig:sub3}
    \end{subfigure}
    \hfill
    \begin{subfigure}[t]{0.48\textwidth}
        \centering
        \includegraphics[width=0.95\textwidth]{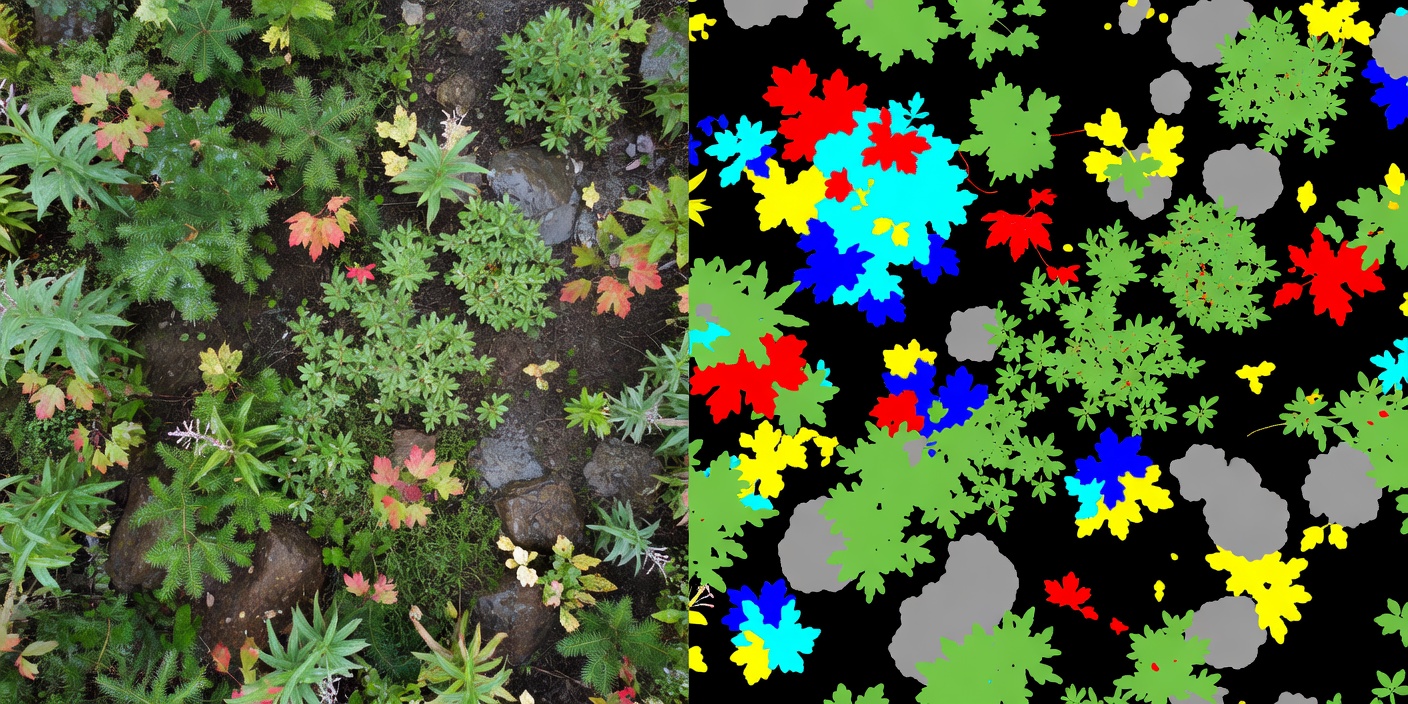}
        \caption{Grok Imagine: The mask colours do not correspond to the actual classes, and some colours were not requested in the prompt.}
        \label{fig:sub4}
    \end{subfigure}
    \caption{Comparison with mainstream image generation models.}
    \label{fig:comparison_other_image_generators}
\end{figure}

Similarly to \citet{gabeur_image_2026}, we noticed the excellent semantic segmentation capabilities of the Nano Banana Pro model, released in November 2025. 
This motivated us to compare Nano Banana Pro with other major commercial image generation models using the same prompt as in Figure~\ref{fig:generation_example_spring}. As shown in Figure~\ref{fig:comparison_other_image_generators}, the outputs of these models are unsatisfactory compared with Nano Banana Pro. Moreover, even when using the same prompt, Nano Banana Pro produces different outputs, which increases diversity during generation.

\newpage
\section{Prompt Generation Failure Cases}
\label{sec:generation_failure_cases}

\begin{figure}[htbp]
    \centering
    \begin{subfigure}[t]{0.48\textwidth}
        \centering
        \includegraphics[width=0.95\textwidth]{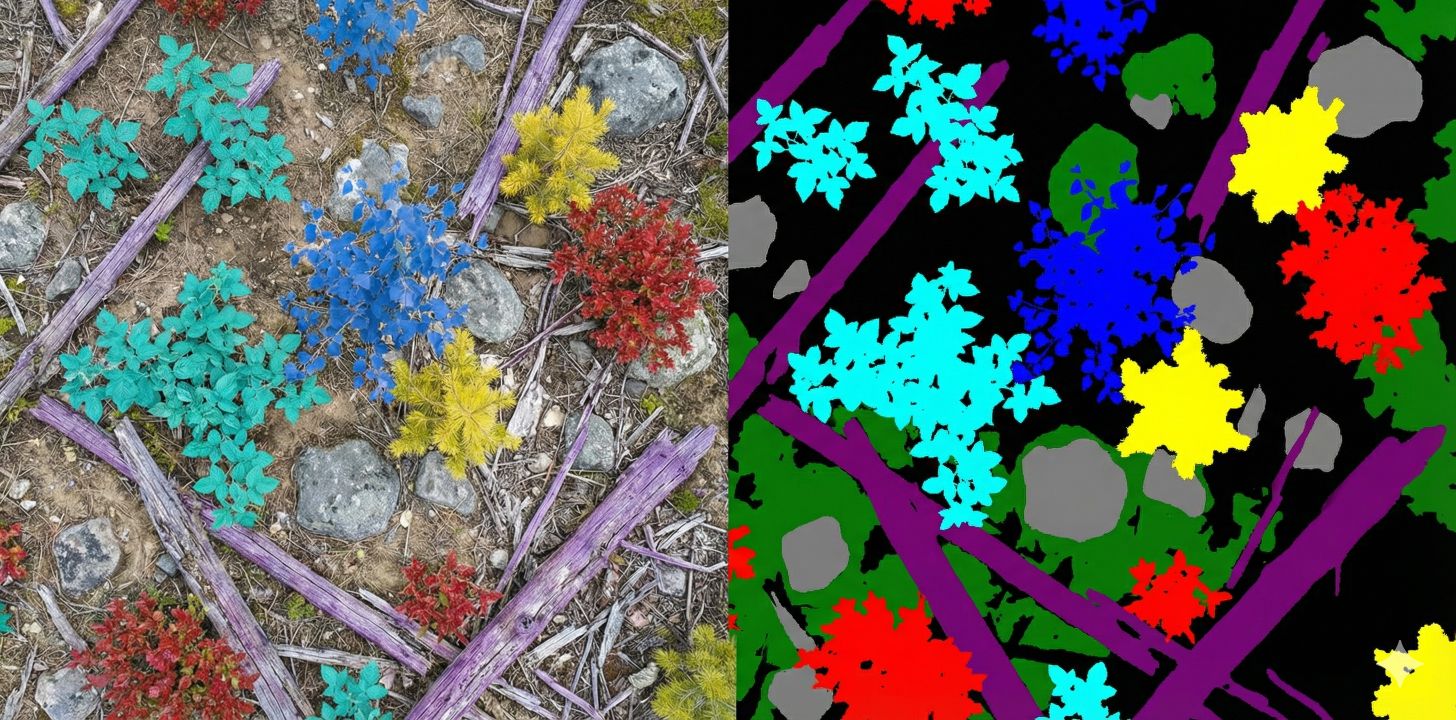}
        \caption{Mask colours leaking into the photorealistic image.}
        \label{fig:sub1}
    \end{subfigure}
    \hfill
    \begin{subfigure}[t]{0.48\textwidth}
        \centering
        \includegraphics[width=0.95\textwidth]{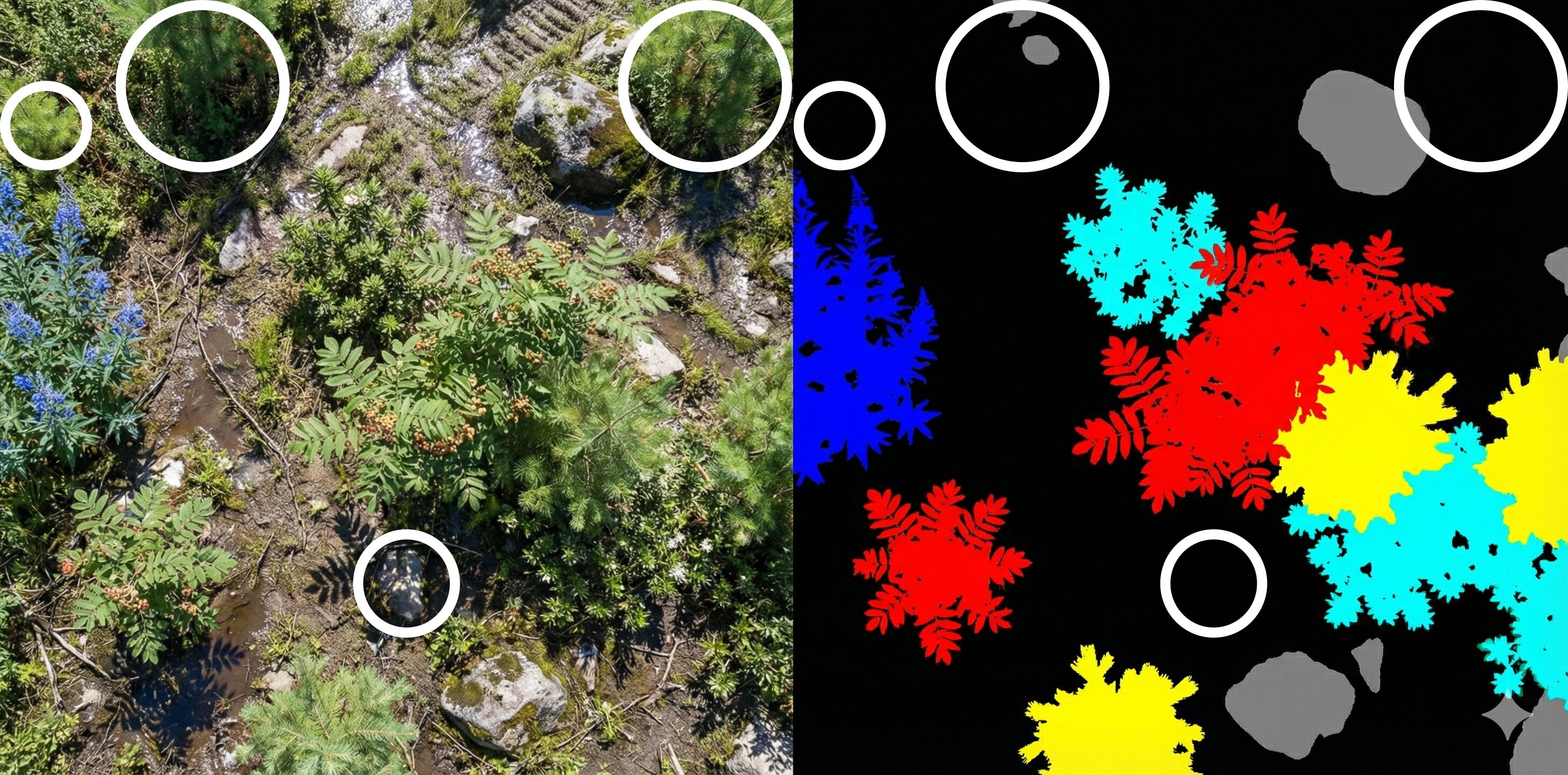}
        \caption{Missing masks, indicated by the white circle.}
        \label{fig:sub2}
    \end{subfigure}
    \begin{subfigure}[t]{0.48\textwidth}
        \centering
        \includegraphics[width=0.95\textwidth]{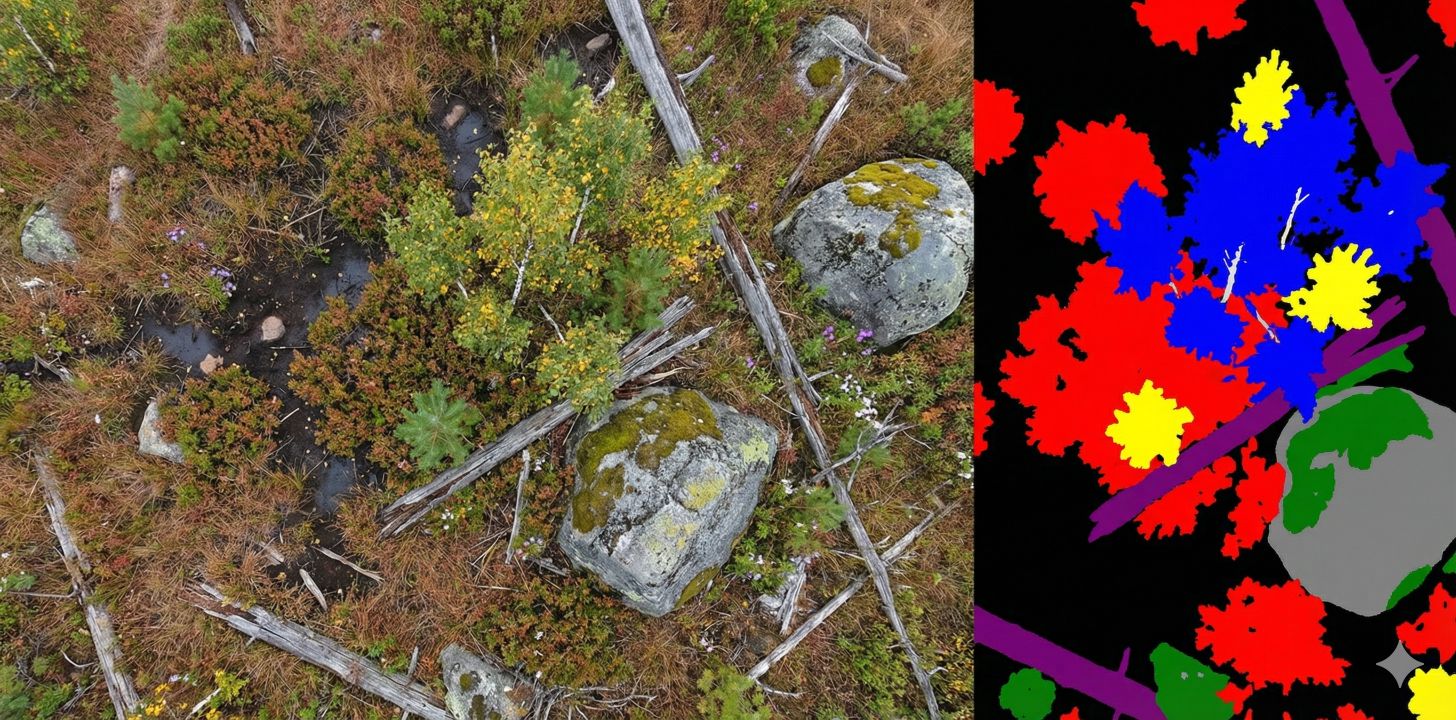}
        \caption{Photorealistic image and mask image having different sizes.}
        \label{fig:sub3}
    \end{subfigure}
    \hfill
    \begin{subfigure}[t]{0.48\textwidth}
        \centering
        \includegraphics[width=0.95\textwidth]{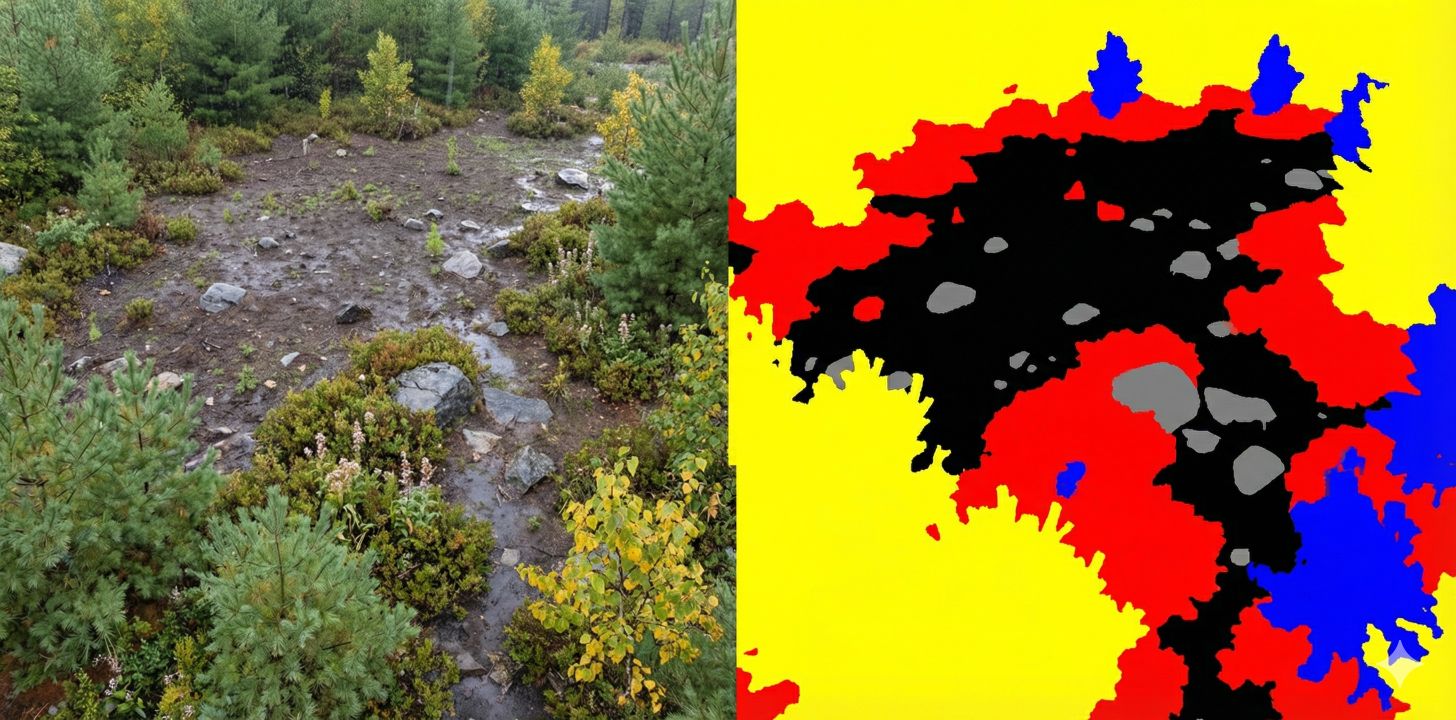}
        \caption{Incorrect viewpoint (not a \ac{UAV} top-view perspective) and slight misalignment of the border between the photorealistic image and corresponding mask image.}
        \label{fig:sub4}
    \end{subfigure}
    \caption{Typical failure cases for image generation.
    See sub-plot captions for the description of each failure.}
    \label{fig:AiGenFailureCases}
\end{figure}

Although the prompt generation with the Nano Banana Pro model has a high success rate, the images and masks generated can still be of lesser quality, hence the importance of doing a visual inspection before adding to the \GenAiDataset dataset.
Some failure cases are displayed in \autoref{fig:AiGenFailureCases}.

\newpage
\section{Zero-Shot Segmentation Prompt}
\label{sec:zeroshot_prompt}

\begin{figure}[!h]
    \centering
    \includegraphics[width=0.95\textwidth]{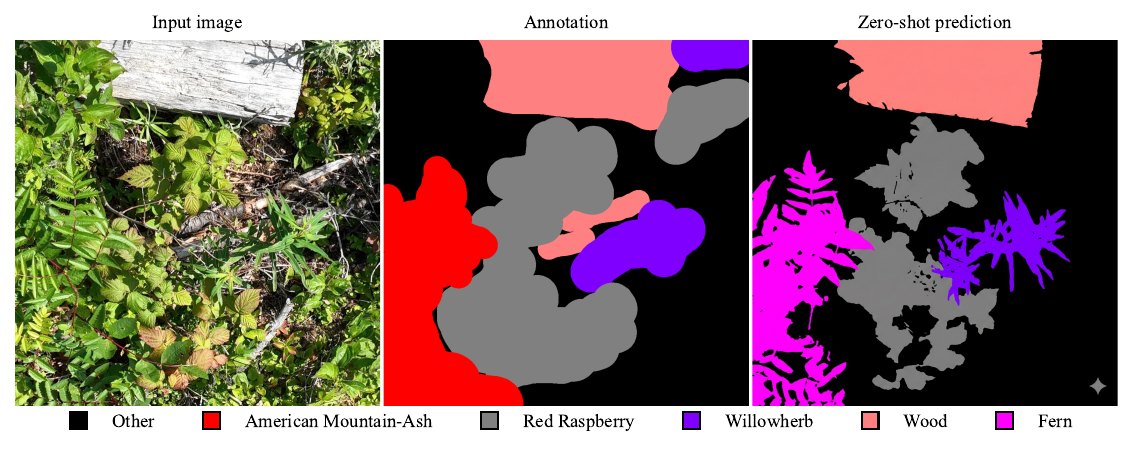}
    \caption{Zero-shot mask generation from this input prompt: }
    \begin{Verbatim}[breaklines=true, breakanywhere=true,
      breaksymbolleft={},
      breaksymbolright={}]
    Attached is a top-view 1024x1024 pixels image taken in a forest regeneration zone in the Quebec province.
    Create a semantic segmentation mask of the image, with the following classes. Next to each class is the colour of the corresponding mask. Make sure the mask is a square, the same size as the image.
    American Mountain-Ash: (255, 0, 0)
    Other: (0, 0, 0)
    Bog Labrador Tea: (0, 255, 0)
    Boulder: (0, 0, 255)
    Canada Yew: (255, 255, 0)
    Fern: (255, 0, 255)
    Fir: (0, 255, 255)
    Fire Cherry: (128, 0, 0)
    Lowbush Blueberry: (0, 128, 0)
    Moss: (0, 0, 128)
    Mountain Maple: (128, 128, 0)
    Paper Birch: (128, 0, 128)
    Pine: (0, 128, 128)
    Red Maple: (192, 192, 192)
    Red Raspberry: (128, 128, 128)
    Sedge: (255, 128, 0)
    Serviceberry: (255, 0, 128)
    Sheep Laurel: (255, 255, 128)
    Spruce: (0, 255, 128)
    Trembling Aspen: (0, 128, 255)
    Willowherb: (128, 0, 255)
    Wood: (255, 128, 128)
    Yellow Birch: (128, 255, 255)
    \end{Verbatim}
    \label{fig:zero_shot_fig}
\end{figure}

The zero-shot semantic segmentation experiment, as shown in \autoref{tab:zero_shot_segmentation}, used a variety of prompts, with different inputs of label lists, as well as using the pseudo-label as an input mask to guide the prediction.
The prompt used for mask generation, when using the full label list as input without the pseudo-label, is presented in \autoref{fig:zero_shot_fig}, alongside an example of prediction and annotation for an image in the dataset.

\end{document}